\pdfoutput=1
\documentclass{article}
\usepackage{iclr2027_conference,times}

\usepackage[utf8]{inputenc}
\usepackage[T1]{fontenc}
\usepackage{url}
\usepackage{booktabs}
\usepackage{amsfonts}
\usepackage{nicefrac}
\usepackage{microtype}
\usepackage{xcolor}

\usepackage{amsmath}
\usepackage{amssymb}
\usepackage{mathtools}
\usepackage{amsthm}
\usepackage{multirow}

\usepackage{graphicx}
\usepackage{wrapfig}
\usepackage{needspace}
\usepackage{algorithm}
\usepackage{algorithmic}

\usepackage{tabularx}
\usepackage{array}

\usepackage{hyperref}

\title{CueRator: Agentic Search for Symbolic Rules to Adapt Frozen Multimodal Encoders}

\author{Sunchan Park\thanks{Equal contribution.}\quad Beomkwon Cho\footnotemark[1]\quad Kyeongbo Kong\thanks{Corresponding author.} \\
Pusan National University \\
\texttt{\{sunchanpark, chobk1, kbkong\}@pusan.ac.kr}}

\iclrfinalcopy

\begin{document}
\pagestyle{plain}

\maketitle

\begin{abstract}
\label{abs}
Large language model agents have been used to search over symbolic structures such as programs and equations. We propose CueRator, an agentic framework for policy-aware decision-rule discovery, which adapts frozen contrastive multimodal encoders by searching for the decision rule that converts their cross-modal similarities into predictions. We validate it on open-vocabulary audio-visual event perception, where existing methods involve a trade-off between adaptivity and generalization to unseen categories: trained modules adapt at the cost of generalization, and fixed rules the reverse. The framework pairs a symbolic formulation for generalization with a lightweight policy that predicts its parameters per video for adaptivity. A report-guided multi-agent loop discovers the formulation offline, evaluating each candidate on its expressive ceiling and on whether a trained policy can realize it. On OV-AVEBench, CueRator raises the total average from 57.8 to 60.2 and unseen-category performance from 55.8 to 59.9 over the best existing method, reducing the seen--unseen gap from 7.1 to 1.2. Ablations attribute the gains to both the formulation and the policy and show that both feedback signals are necessary for effective search. CueRator also improves over the respective baselines on two further audio-visual event perception tasks, and the discovered rule remains competitive across encoders with only the policy retrained. Code is available at \url{https://github.com/cvsp-lab/cuerator}.
\end{abstract}
\section{Introduction}
\label{int}

Large language models (LLMs) are increasingly used not only to answer questions but to search over symbolic structures such as programs and equations: to propose candidates in a symbolic form, evaluate them automatically, and refine them over successive iterations. This form of agentic search has produced programs that yield new constructions in combinatorics~\citep{funsearch, alphaevolve} and equations discovered from scientific data~\citep{llmsr, srscientist}. The approach applies when the object of interest can be written as a program and scored automatically. Under these conditions, the model's reasoning about structure can be coupled to execution feedback, and search becomes possible in spaces that are difficult to explore by hand.

Contrastive multimodal encoders such as CLIP, CLAP, and ImageBind~\citep{clip, clap, imagebind} embed inputs from different modalities into a shared space. Categories are specified in text, and an input is scored by its similarity to their descriptions in that space. This makes the encoders usable in open-vocabulary settings: even categories that do not appear in the downstream training data can be recognized. To adapt them to a task while keeping this ability, existing methods typically either feed the frozen embeddings into modules trained for the task or convert the similarity scores into predictions with a decision rule. Such a rule is the kind of object described above: it can be written as a program, and its predictions can be scored against labeled data.

We study this rule in audio-visual event perception, the temporal localization and classification of events from audio and visual cues, in the open-vocabulary setting where some test categories are unseen during training. Existing methods either train temporal modules on seen categories or apply a fixed agreement rule or dynamic thresholds over the similarity scores~\citep{ov-avel, av2a}. Trained modules offer adaptivity at the cost of generalization, whereas fixed rules offer generalization at the cost of adaptivity. This suggests a symbolic rule whose parameters are predicted per input: the rule provides generalization, and the per-input parameters provide adaptivity.

Following this direction, we cast the problem as \emph{policy-aware decision-rule discovery}: the rule takes the form of a symbolic formulation, which serves as an interpretable scaffold, and its parameters are predicted by a neural policy. The formulation is not written by hand but found by agentic search, with LLM agents proposing candidates in executable form and refining them from evaluation on labeled data. Since the formulation and the policy operate together, candidates are evaluated on both how well they can perform and whether the policy can learn to predict their parameters.

\begin{figure*}[t]
  \begin{center}
    \centerline{\includegraphics[width=\textwidth]{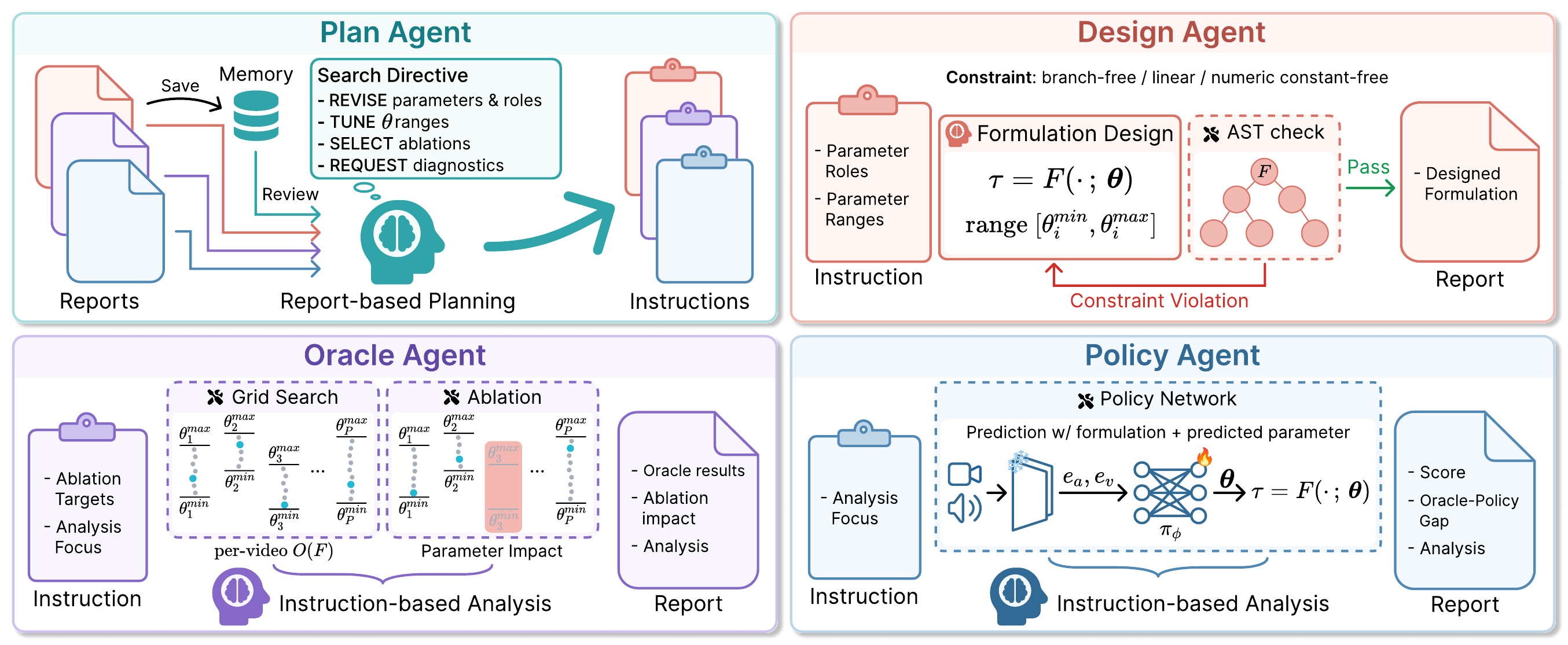}}
\caption{
\textbf{Overview of CueRator's report-guided agentic search.}
At each iteration, the Plan Agent sets the search direction from accumulated reports, the Design Agent proposes an executable formulation verified against formal constraints, and the Oracle and Policy Agents evaluate the candidate on how well it can perform and whether its parameters can be learned. The Oracle Agent estimates the expressive ceiling through per-video parameter search and ablations, while a policy network is trained for the candidate and diagnosed by the Policy Agent. All reports are appended to memory for the next iteration; no LLM agent is used at test time.
}
    \label{fig:overview}
  \end{center}
\end{figure*}

In this paper, we propose \textbf{CueRator}, an agentic framework for policy-aware decision-rule discovery, in which a symbolic formulation $F(\cdot;\theta)$ maps the similarity cues to audio and visual dynamic thresholds and a lightweight policy network $\pi_\phi$ predicts its parameters $\theta$ per video. The formulation is found by a report-guided multi-agent loop (Fig.~\ref{fig:overview}), in which agents exchange written reports as well as scores: a \emph{Plan Agent} curates past reports into a search direction, a \emph{Design Agent} proposes executable formulations under formal constraints, an \emph{Oracle Agent} estimates each candidate's expressive ceiling through per-video parameter search, and a \emph{Policy Agent} diagnoses a policy trained on the candidate. LLM agents are used only in this offline search; at test time, inference requires only one policy-network forward pass and deterministic thresholding.

We evaluate CueRator on OV-AVEBench~\citep{ov-avel}, the benchmark for open-vocabulary audio-visual event localization (OV-AVEL), against existing training-free, fine-tuning, and dynamic-thresholding methods. CueRator, which trains only a lightweight policy network, achieves the highest overall performance. Ablations attribute the gains to both components, the discovered formulation over hand-designed rules and the per-video policy over globally fixed parameters, and show that the search requires diagnostic feedback from both the Oracle and Policy Agents to find formulations that remain usable after policy instantiation. The discovery framework, adapted to two adjacent audio-visual event perception tasks, also improves over the respective baselines.

\needspace{5\baselineskip}
Our contributions are threefold:
\begin{itemize}
    \item We propose \textbf{CueRator}, an agentic framework for \emph{policy-aware decision-rule discovery}, which adapts frozen contrastive multimodal encoders with a symbolic formulation for generalization and a neural policy predicting its parameters per input for adaptivity.

    \item We design its report-guided multi-agent loop, in which Plan, Design, Oracle, and Policy Agents propose, evaluate, and diagnose executable formulations under formal constraints, judging each candidate on both its expressive ceiling and policy learnability.

    \item We validate CueRator on three open-vocabulary audio-visual event perception tasks: it improves over the respective baselines on OV-AVEL and two adjacent tasks without encoder fine-tuning, and its discovered rule remains competitive across encoders.
\end{itemize}
\section{Related Work}
\label{rel}

\paragraph{LLM-guided symbolic search.}
Large language models have been used to propose symbolic candidates, evaluate them automatically, and refine them over iterations. This loop has produced new constructions and algorithms in mathematics~\citep{funsearch, alphaevolve}, scientific equations~\citep{llmsr, sr1, sr3, srscientist}, constitutive laws and molecules~\citep{sr2}, heuristics for combinatorial optimization~\citep{eoh, reevo}, reward functions for robot learning~\citep{eureka}, preference-optimization losses~\citep{discopop}, candidate materials~\citep{llema}, and point sets for numerical integration~\citep{qmcevolve}. CueRator instead searches for a decision rule that adapts frozen multimodal encoders, evaluating candidates on both expressiveness and policy learnability (Section~\ref{sec:overview}).

\paragraph{Audio-visual event perception.}
AVEL localizes temporal segments where an event is both audible and visible~\citep{avel}; closed-vocabulary methods improve fusion~\citep{avfusion4, avfusion3, psp}, background discrimination~\citep{back1, back2}, and boundary localization~\citep{mmpyramid, tempbound1}, and related tasks include audio-visual video parsing (AVVP)~\citep{avvp, avvp2, avvp4, avvp5} and dense audio-visual event localization (DAVEL)~\citep{davel, avfusion5, davel2}.
OV-AVEBench~\citep{ov-avel} extends AVEL to the open-vocabulary setting. Existing methods either train temporal modules on frozen embeddings, or convert cross-modal similarities into predictions with agreement rules or dynamic thresholds~\citep{ov-avel, av2a}. CueRator differs by searching for the decision rule itself and instantiating its parameters per video.
\section{Method}
\label{met}
\label{sec:method}

\subsection{Problem Setup}
\label{sec:problem-setup}

\begin{wrapfigure}{r}{0.5\textwidth}
  \centering
  \vspace{-10pt}
  \includegraphics[width=0.5\textwidth]{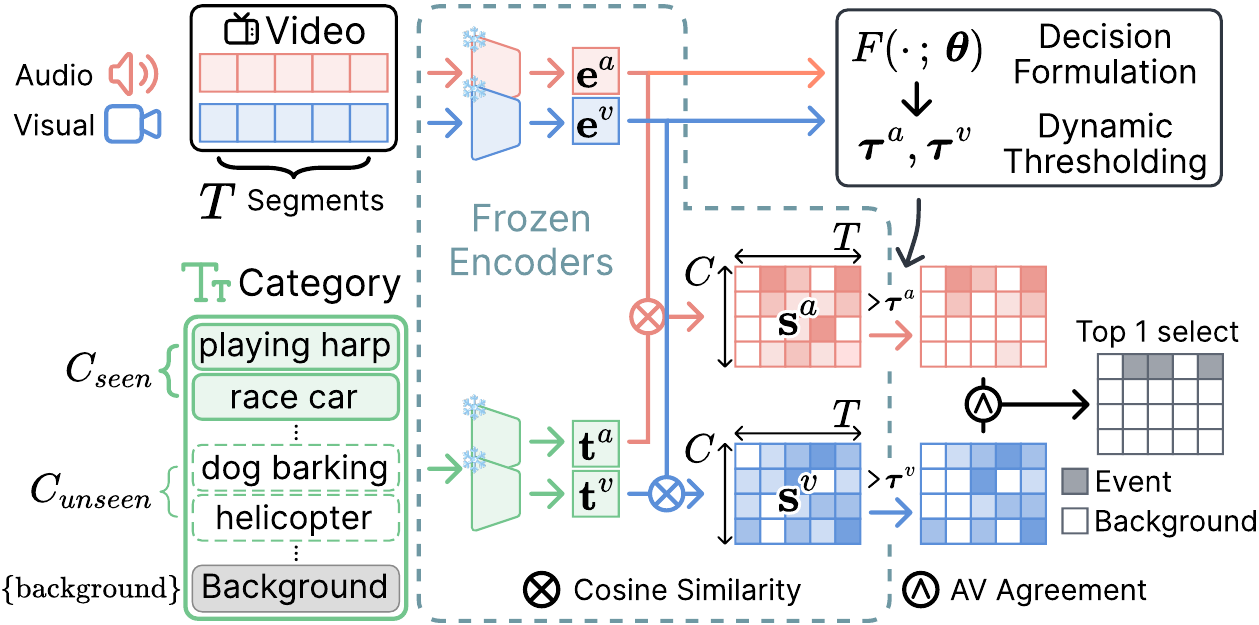}
  \caption{Overview of the OV-AVEL pipeline.}
  \label{fig:setup}
  \vspace{-10pt}
\end{wrapfigure}

As illustrated in Fig.~\ref{fig:setup}, OV-AVEL~\citep{ov-avel} divides each video into $T$ temporal segments and assigns to every segment a label from $C_{\text{seen}} \cup C_{\text{unseen}} \cup \{\text{background}\}$. Models are trained on segments whose foreground labels belong to $C_{\text{seen}}$, but at test time the ground-truth label of a segment may also belong to $C_{\text{unseen}}$, requiring generalization beyond the training categories. The pipeline consists of frozen encoders that extract audio-visual similarity scores, which we call cues, and a decision rule that maps those cues to per-segment predictions, whose executable form we call a decision \emph{formulation}.

\textbf{Frozen Encoders.} We extract all representations from frozen pretrained encoders. For a video with $T$ segments and a category set of size $C$, we obtain per-segment audio and visual embeddings $\mathbf{e}^a, \mathbf{e}^v \in \mathbb{R}^{T \times D}$ and per-category text embeddings $\mathbf{t}^a, \mathbf{t}^v \in \mathbb{R}^{C \times D}$, where each text embedding is aligned to its corresponding modality space. When the encoder maps all modalities into a common space (e.g., ImageBind), $\mathbf{t}^a$ and $\mathbf{t}^v$ are identical. We denote by $s_{tc}^{m}$ the cosine similarity between the embedding of modality $m \in \{a, v\}$ at segment $t$ and the text embedding of category $c$, yielding similarity matrices $\mathbf{s}^a, \mathbf{s}^v \in \mathbb{R}^{T \times C}$.

\textbf{Formulation Interface.} A decision formulation \(F(\cdot;\boldsymbol{\theta})\) takes frozen embeddings and similarity matrices \((\mathbf{s}^a,\mathbf{s}^v)\) as input, composes them through deterministic primitives, and produces per-segment, per-category thresholds \(\boldsymbol{\tau}^a,\boldsymbol{\tau}^v \in \mathbb{R}^{T \times C}\). A segment $t$ first filters out categories below the threshold in each modality, then selects the top-scoring category per modality among the remaining ones. The segment is assigned that category if both modalities agree, and labeled background otherwise. This decision procedure is fixed across all candidate formulations; only the threshold-producing function $F$ varies during the agentic search. The parameter vector $\boldsymbol{\theta} \in \mathbb{R}^P$, where $P$ is the number of parameters, controls the behavior of $F$.

\subsection{CueRator Overview}
\label{sec:overview}
CueRator adapts frozen encoders at the decision layer. It separates prediction into two components: a symbolic formulation \(F(\cdot;\boldsymbol{\theta})\), which defines how the similarity cues are converted into audio and visual thresholds, and a lightweight policy network \(\pi_\phi\), which predicts the formulation parameters \(\boldsymbol{\theta}\) for each video from frozen audio-visual embeddings \((\mathbf{e}^a,\mathbf{e}^v)\). The formulation provides the decision-rule scaffold, while the policy supplies video-specific adaptation within that scaffold.

\begin{algorithm}[t]
\caption{Policy-aware decision-rule discovery (one session)}
\label{alg:search}
\begin{algorithmic}[1]
\REQUIRE $(\mathbf{e}^a,\mathbf{e}^v,\mathbf{s}^a,\mathbf{s}^v)$ of training and validation videos; iterations $N$; memory $\mathcal{M}\leftarrow\emptyset$
\FOR{$i=1,\dots,N$}
  \STATE \emph{Plan Agent}: read $\mathcal{M}$, write directives $d_i$
  \STATE \emph{Design Agent}: propose formulation $F_i$ with parameter ranges $\Theta_i$ under formal constraints
  \STATE Per-video oracle search of $\boldsymbol{\theta}\in\Theta_i$ $\Rightarrow$ oracle ceiling $\mathcal{O}(F_i)$; \emph{Oracle Agent}: report $R^{\mathrm{O}}_i$
  \STATE Train $\pi_{\phi_i}$ to predict $\boldsymbol{\theta}$ $\Rightarrow$ validation score $\mathcal{V}(F_i,\pi_{\phi_i})$; \emph{Policy Agent}: report $R^{\mathrm{P}}_i$
  \STATE $\mathcal{M}\leftarrow\mathcal{M}\cup\{d_i,F_i,\Theta_i,\mathcal{O}(F_i),\mathcal{V}(F_i,\pi_{\phi_i}),R^{\mathrm{O}}_i,R^{\mathrm{P}}_i\}$
\ENDFOR
\RETURN $(F^\star,\pi_\phi^\star)$ with the highest $\mathcal{V}$
\end{algorithmic}
\end{algorithm}

Fig.~\ref{fig:overview} and Algorithm~\ref{alg:search} summarize the offline search; LLM agents are used only in this search, never at test time. At each iteration, the \emph{Plan Agent} sets search directives from the accumulated memory, the \emph{Design Agent} proposes an executable formulation and parameter ranges under formal constraints, the \emph{Oracle Agent} evaluates the candidate's expressive capacity through per-video parameter search and requested ablations, and a policy trained for the same candidate is diagnosed by the \emph{Policy Agent}. All reports are appended to memory, so that each iteration's directives are grounded in the full history of candidates rather than the most recent one alone.

This makes CueRator policy-aware, but not a full co-optimization procedure. For each candidate formulation \(F_i\), CueRator records two complementary signals:
\[
q_i^{\mathrm{oracle}} = \mathcal{O}(F_i),
\qquad
q_i^{\mathrm{policy}} = \mathcal{V}(F_i,\pi_{\phi_i}),
\]
where \(\mathcal{O}(F_i)\) measures the best attainable performance under ideal per-video parameters, and \(\mathcal{V}(F_i,\pi_{\phi_i})\) measures validation performance after training a policy to instantiate the formulation. Throughout the paper, we refer to $\mathcal{O}(F_i)$ as the formulation's \emph{oracle ceiling}---the performance of $F_i$ when $\boldsymbol{\theta}$ is set optimally for each video, an upper bound on what the rule's structure can express---and to \emph{learnability} as whether the optimal $\boldsymbol{\theta}$ is predictable from the video's embeddings, measured as the gap between the oracle ceiling and policy-instantiated performance. The final formulation--policy pair is selected by validation performance, while the test split is held out from all agents and used only for final evaluation in Section~\ref{exp}. Appendix~\ref{app:aged} provides an input--operation--output summary of the four agents and representative agent-visible prompts and reports.

\subsection{Plan Agent}
\label{sec:plan_agent}

\textbf{Search trajectory analysis.} The Plan Agent begins each iteration by reviewing memory to assess the current state of the search. Memory stores per-iteration metrics, one-line agent summaries, and full per-agent reports accessed on demand. For example, a widening gap between the oracle ceiling and validation policy performance indicates a learnability problem: the policy fails to predict optimal parameters from context. In another case, when neither oracle nor validation performance improves, the formulation itself is the bottleneck, requiring an axis shift rather than incremental refinement.

\textbf{Directive construction.} Based on this diagnosis, the Plan Agent produces three directives: a natural-language strategy memo directed to the Design Agent, an evaluation protocol specifying parameter ablations paired with analysis intents for the Oracle Agent, and an analysis directive for the Policy Agent. The strategy memo assigns each parameter a conceptual role and an operating range. When the Oracle or Policy reports identify a parameter as inactive, the Plan Agent reallocates it to capture a different aspect of the audio-visual similarity structure. The evaluation protocol and analysis directive are grounded in the same parameter definitions as the strategy memo, so that oracle expressivity and policy learnability analyses test the same hypotheses.

\subsection{Design Agent}
\label{sec:design_agent}

\textbf{Formulation proposal.} Given the Plan Agent's directive, the Design Agent proposes a candidate formulation as a Python function together with a range $[\theta_i^{\min}, \theta_i^{\max}]$ for each parameter, jointly denoted $\Theta=\prod_i[\theta_i^{\min},\theta_i^{\max}]$, typically including zero so that a term can be deactivated. These ranges define both the search space for oracle evaluation and the output bounds for the policy network. The agent receives the interface specification---the available inputs are frozen embeddings and similarity matrices \((\mathbf{s}^a,\mathbf{s}^v)\), from which the agent defines deterministic primitives.

\textbf{Formal constraints.} Each candidate formulation must satisfy a set of structural constraints that together ensure the resulting threshold equations are interpretable as closed-form expressions. The threshold for each modality takes a fixed linear form: a bias plus a system-specified number of weighted terms. Each term is defined as a symbolic function over the available audio-visual inputs. To avoid non-linear parameter interactions that destabilize both Sobol oracle search and policy gradient learning, each parameter may only appear as an outer weight or bias in the threshold expression, never as an inner argument to a term. Numeric constants within the formulation are restricted to a small predefined set $\{0, \pm 0.5, \pm 1, \pm 2\}$ plus a numerical stability epsilon, with small integer literals ($|n|\le 16$) allowed only for indexing and shape arguments, preventing the Design Agent from inserting hidden hyperparameters.

\textbf{Constraint verification.} These constraints are specified as natural-language instructions in the prompt and verified via abstract syntax tree (AST) parsing, covering function count, per-function statement budget, numeric literals, inner-parameter usage, and control flow. The prohibition on control flow — no conditionals, loops, or comprehensions — requires all logic to be expressed as closed-form expressions, preserving the interpretability of the formulation. Candidates that fail any check are returned to the Design Agent with diagnostic feedback for revision.

\subsection{Oracle Agent}
\label{sec:oracle_agent}

\textbf{Oracle ceiling.} For each candidate formulation $F(\cdot\,;\,\boldsymbol{\theta})$, the Oracle Agent performs a per-video parameter search over the ranges $[\theta_i^{\min}, \theta_i^{\max}]$ defined by the Design Agent, using a Sobol sequence to provide more uniform coverage. For each training video $n$, the optimal parameters are obtained as $\boldsymbol{\theta}^\star(n) = \arg\max_{\boldsymbol{\theta}} \mathcal{R}(F, \boldsymbol{\theta}; n)$, where $\mathcal{R}$ is a per-video performance metric. The oracle ceiling is then defined as:
\[
\mathcal{O}(F) = \mathbb{E}_n \big[ \mathcal{R}(F, \boldsymbol{\theta}^\star(n); n) \big],
\]
which measures the formulation's structural expressiveness independently of how well a policy can predict the optimal parameters in practice.

\textbf{Parameter ablation.} Following the evaluation protocol issued by the Plan Agent, each designated parameter subset is independently zeroed out and the remaining parameters are re-optimized via a separate Sobol grid to obtain per-video ablated optimal parameters $\boldsymbol{\theta}^\star_{\text{abl}}(n)$. The per-sample marginal contribution of each designated subset is then defined as:
\[
\Delta(n) = \max\!\big(0,\, \mathcal{R}(F, \boldsymbol{\theta}^\star(n); n) - \mathcal{R}(F, \boldsymbol{\theta}^\star_{\text{abl}}(n); n)\big),
\]
where the $\max(0,\cdot)$ clipping handles finite Sobol sampling noise: the ablated search space is a subset of the full one, so the inequality $\mathcal{R}(F, \boldsymbol{\theta}^\star_{\text{abl}}(n); n) \leq \mathcal{R}(F, \boldsymbol{\theta}^\star(n); n)$ holds in principle, but empirical optima from a finite grid may slightly violate it, yielding a distribution over training videos that quantifies how much each parameter group contributes to the oracle ceiling.

\textbf{Analysis and report.} Beyond the predefined oracle evaluation flow, the Oracle Agent writes and executes custom scripts against the raw per-sample data to conduct analyses at three levels. First, standard analyses covering parameter distribution statistics, modality balance, and per-ablation marginal patterns are conducted. Second, targeted analyses follow the Plan Agent's directives, examining specific hypotheses about parameter roles or ablation patterns. Third, the agent selects additional analyses based on what the results suggest, such as per-category marginal breakdowns or cross-ablation correlation. All findings are compiled into a natural-language report with a one-line summary, both appended to memory for the Plan Agent at the next iteration.

\subsection{Policy Agent}
\label{sec:policy_agent}

\textbf{Policy network.} To provide video-specific parameter adaptation within the candidate formulation, a lightweight policy network $\pi_\phi$ is trained for each candidate; it takes a sequence of per-segment frozen embeddings $(\mathbf{e}^a, \mathbf{e}^v)$ as input and outputs a single parameter vector $\boldsymbol{\theta}$ per video. We implement $\pi_\phi$ with modality-specific self-attention layers followed by cross-attention between the two modalities. The resulting representations are mean-pooled over segments, concatenated, and passed to a linear head to produce a raw output $\mathbf{z} \in \mathbb{R}^P$. Each $z_i$ is then squashed into the per-parameter range $[\theta_i^{\min}, \theta_i^{\max}]$ defined with the formulation via the logistic sigmoid:
\[
\theta_i = \theta_i^{\min} + \sigma(z_i)\cdot(\theta_i^{\max} - \theta_i^{\min}).
\]
These ranges, iteratively refined across iterations based on accumulated oracle and policy reports, serve as a hard prior on each parameter's operating region.

\textbf{Policy gradient training.} We cast per-video parameter prediction as a contextual bandit with state $(\mathbf{e}^a, \mathbf{e}^v)$, action $\boldsymbol{\theta}$, and reward $\mathcal{R}(F, \boldsymbol{\theta}; n)$. We train $\pi_\phi$ with policy gradient, as the decision procedure of Section~\ref{sec:problem-setup} makes the reward non-differentiable and per-video oracle parameters do not provide a unique regression target. We use leave-one-out REINFORCE~\citep{leaveoneout} with a Gaussian policy (mean $\mathbf{z}$, fixed standard deviation), drawing $K$ actions per video and using the mean reward of the remaining $K-1$ as an unbiased per-context baseline to reduce variance. At inference, the mean $\mathbf{z}$ is used directly. After training for a fixed number of epochs, the checkpoint with the highest validation performance is selected.

\textbf{Analysis and report.} The selected checkpoint is applied to the validation set to produce per-sample outputs including predicted parameter vectors, per-modality frame-level predictions, and performance metrics, all stored as raw data. The Policy Agent then writes and executes custom scripts against this data following the same three-level structure as the Oracle Agent — standard, directed, and custom — with a focus on learnability. The findings are then compiled into a report and a one-line summary appended to memory for the next iteration.
\section{Experiments}
\label{exp}

\subsection{Setup}
\label{sec:exp-setup}

\textbf{Evaluation.} We evaluate on OV-AVEBench~\citep{ov-avel}: $24{,}800$ 10-second videos across $67$ categories ($46$ seen, $21$ unseen). Training uses seen-only videos; validation and test include both seen and unseen classes. We use frozen ImageBind~\citep{imagebind} as the shared multimodal encoder for audio, visual, and text modalities. Following~\citep{ov-avel}, we report segment-level accuracy (Acc.), segment-level F1-score (Seg.), event-level F1-score at Intersection over Union (IoU)~$\geq$~0.5 (Eve.), and their average (Avg.). Baselines include the OV-AVEBench~\citep{ov-avel} training-free models (Video-LLaMA2~\citep{videollama}, CLIP\&CLAP~\citep{clip, clap}, ImageBind-TF~\citep{imagebind}), fine-tuning models (CMRA, AVE, PSP, MM-Pyramid, ImageBind-FT~\citep{cmra, avel, psp, mmpyramid, imagebind}), and AV$^2$A~\citep{av2a}.

\textbf{Implementation.} The agentic search runs 3 isolated sessions of 20 iterations each, yielding 60 candidate formulations. All agents use gpt-5.4~\citep{gpt5} in OpenAI Codex (v0.125.0, April 2026). Each formulation defines $P{=}10$ parameters with ranges $[\theta_i^{\min}, \theta_i^{\max}]$. The oracle uses a Sobol sequence of $2^{17}$ $\boldsymbol{\theta}$ samples per video. All $\pi_\phi$ share the same hyperparameters: dim $256$, $4$ heads, FFN $512$, dropout $0.1$, $\sigma{=}0.3$; trained with AdamW~\citep{adamw} (lr $1{\times}10^{-4}$, batch $32$, $20$ epochs) on a single RTX 3090, selecting the best validation Acc.\ checkpoint. The leave-one-out REINFORCE~\citep{leaveoneout} objective draws $K{=}4$ samples with video-level Acc.\ as reward. The best-validation candidate across sessions is taken as $F^\star$ with policy $\pi_\phi^\star$. Each session takes 9.8 hours on average.

\begin{table*}[t]
\centering
\caption{\textbf{Performance comparison on OV-AVEBench.} Baseline scores are from~\citep{ov-avel} except AV$^2$A, which we adapt to ImageBind. Fine-tuning rows are OV-AVEBench reimplementations of~\citep{ov-avel}, which share the same frozen ImageBind representations and differ only in the trainable module on top. AV$^2$A hyperparameters are tuned on a 200-video validation subset following its original protocol (Appendix~\ref{app:expd}). AV$^2$A-Policy retains the AV$^2$A rule but replaces its global parameters with CueRator's per-video policy network (Appendix~\ref{app:expd}). CueRator is the validation-selected candidate from three search sessions (mean $\pm$ std over sessions in Table~\ref{tab:ablation_s1}).}
\vspace{0.5em}
\label{tab:main_result}

\setlength{\tabcolsep}{4pt}
\small

\begin{tabular}{l|cccc|cccc|cccc}
\toprule
& \multicolumn{4}{c}{Seen}
& \multicolumn{4}{c}{Unseen}
& \multicolumn{4}{c}{Total} \\

\cmidrule(lr){2-5}
\cmidrule(lr){6-9}
\cmidrule(lr){10-13}

Method
& Acc. & Seg. & Eve. & \textbf{Avg.}
& Acc. & Seg. & Eve. & \textbf{Avg.}
& Acc. & Seg. & Eve. & \textbf{Avg.} \\

\midrule
\multicolumn{13}{l}{\textit{Training-free}} \\
\hspace{1em}Video-LLaMA2
& 50.1 & 40.6 & 32.0 & 40.9
& 48.5 & 38.5 & 29.0 & 38.6
& 48.9 & 39.1 & 29.8 & 39.3 \\

\hspace{1em}CLIP\&CLAP
& 51.4 & 41.4 & 31.9 & 41.6
& 51.6 & 42.2 & 31.6 & 41.8
& 51.5 & 41.9 & 31.7 & 41.7 \\

\hspace{1em}ImageBind-TF
& 57.5 & 45.0 & 34.0 & 45.5
& 59.8 & 47.3 & 34.0 & 47.0
& 59.2 & 46.7 & 34.0 & 46.6 \\

\midrule
\multicolumn{13}{l}{\textit{Fine-tuning}} \\
\hspace{1em}CMRA
& 65.2 & 58.8 & 54.3 & 59.4
& 36.0 & 31.0 & 26.3 & 31.1
& 44.3 & 38.9 & 34.3 & 39.2 \\

\hspace{1em}AVE
& \textbf{76.6} & 63.6 & 56.0 & 65.4
& 44.6 & 33.2 & 24.0 & 34.0
& 53.8 & 41.9 & 33.2 & 42.9 \\

\hspace{1em}PSP
& 75.4 & 66.8 & 61.0 & 67.7
& 33.7 & 28.2 & 24.2 & 28.7
& 45.6 & 39.3 & 34.7 & 39.9 \\

\hspace{1em}MM-Pyramid
& 76.5 & \textbf{66.9} & \textbf{62.3} & \textbf{68.6}
& 36.8 & 29.0 & 23.8 & 29.9
& 48.4 & 40.2 & 35.2 & 41.2 \\

\hspace{1em}ImageBind-FT
& 72.5 & 61.8 & 54.5 & 62.9
& 64.9 & 55.0 & 47.5 & 55.8
& 67.1 & 56.9 & 49.5 & 57.8 \\

\midrule
\multicolumn{13}{l}{\textit{Dynamic Thresholding}} \\

\hspace{1em}AV$^2$A
& 59.0 & 52.4 & 48.7 & 53.4
& 62.4 & 54.5 & 49.8 & 55.6
& 61.4 & 53.9 & 49.5 & 55.0 \\

\hspace{1em}AV$^2$A-Policy
& 67.0 & 60.2 & 57.0 & 61.4
& 62.7 & 55.6 & 51.2 & 56.5
& 63.9 & 56.9 & \textbf{52.9} & 57.9 \\

\hspace{1em}\textbf{CueRator~(ours)}
& 68.6 & 60.1 & 54.6 & 61.1
& \textbf{68.5} & \textbf{59.1} & \textbf{52.0} & \textbf{59.9}
& \textbf{68.6} & \textbf{59.4} & 52.7 & \textbf{60.2} \\

\bottomrule
\end{tabular}

\end{table*}

\subsection{Quantitative Results}

\textbf{Overall performance.} We organize baselines into three groups: training-free, fine-tuning, and dynamic thresholding methods. Since AV$^2$A has no published ImageBind configuration, we adapt it ourselves and tune its hyperparameters on a validation subset following its original protocol (Appendix~\ref{app:expd}). CueRator achieves the highest Total Avg.\ of $60.2$ among all baselines (Table~\ref{tab:main_result}). This candidate is selected by validation performance from three search sessions, whose test Total Avg.\ averages $59.4 \pm 0.9$ (Table~\ref{tab:ablation_s1}). Within the dynamic thresholding group, CueRator improves over AV$^2$A by $+5.2$ Avg., indicating that per-video adaptive parameterization captures structure that a globally tuned rule cannot. CueRator also outperforms the best fine-tuning baseline, ImageBind-FT, by $+2.4$ Avg.\ while training only a lightweight policy network.

\textbf{Formulation and policy contributions.} AV$^2$A-Policy isolates the two components by pairing AV$^2$A's manually designed rule with CueRator's policy network and training protocol. With the rule fixed, per-video adaptation improves over AV$^2$A by $+2.9$ Avg., but the gain is concentrated on seen categories ($+8.0$ Seen, $+0.9$ Unseen). With the policy fixed, the discovered formulation adds a further $+2.3$ Avg.; the gain comes from Unseen ($+3.4$) with Seen nearly unchanged ($-0.3$), reducing the seen--unseen gap from $4.9$ to $1.2$. Neither component alone explains the improvement: the policy provides video-specific adaptation, while the discovered formulation enables this adaptation to generalize to unseen categories. Qualitative examples of the per-video thresholds are shown in Appendix~\ref{app:qual}.

\textbf{Seen--unseen balance.} Training-free methods tend to show relatively balanced Seen and Unseen
performance, though their absolute accuracy remains limited. Fine-tuning methods (AVE, PSP, MM-Pyramid) achieve strong Seen performance but suffer severe degradation on Unseen categories, with gaps exceeding $30$ points---a clear sign of seen-class overfitting. ImageBind-FT, which freezes the encoder and trains only additional temporal layers, narrows the gap to $7.1$ points. Dynamic thresholding methods achieve higher overall performance than training-free baselines while maintaining better Seen--Unseen balance than fine-tuning methods. CueRator reaches the highest Total Avg.\ with a Seen--Unseen gap of $1.2$ points, combining the balance of fixed rules with a higher overall level than trained modules.

\needspace{14\baselineskip}
\subsection{Ablation Study}

\begin{wraptable}{r}{0.55\textwidth}
\vspace{-22pt}
\centering
\caption{\textbf{Ablation study.} Mean $\pm$ std.\ over three sessions. \emph{Oracle}: accuracy with per-video oracle parameters on the training set. \emph{Test}: test-set Acc.\ and Avg. For global threshold, Oracle uses a single global $\boldsymbol{\theta}$. The validation-selected candidate attains 60.2 (Table~\ref{tab:main_result}).}
\vspace{0.5em}
\label{tab:ablation_s1}
\small
\setlength{\tabcolsep}{4.5pt}
\begin{tabular}{lccc}
\toprule
\multirow{2}{*}[-0.5ex]{Variant}
  & \multicolumn{1}{c}{Oracle}
  & \multicolumn{2}{c}{Test} \\
\cmidrule(lr){2-2}\cmidrule(lr){3-4}
 & Acc. & Acc. & Avg. \\
\midrule
CueRator (full)      & 88.7 $\pm$ 0.9 & \textbf{68.3 $\pm$ 0.8} & \textbf{59.4 $\pm$ 0.9} \\
\midrule
w/o policy agent & \textbf{91.5 $\pm$ 2.3} & 66.7 $\pm$ 0.9 & 56.3 $\pm$ 1.7 \\
w/o oracle agent & 87.3 $\pm$ 5.0 & 66.0 $\pm$ 0.2 & 55.3 $\pm$ 0.5 \\
score-only loop  & 88.4 $\pm$ 1.3 & 64.1 $\pm$ 1.1 & 53.6 $\pm$ 0.7 \\
no feedback      & 86.0 $\pm$ 3.5 & 62.2 $\pm$ 2.2 & 51.8 $\pm$ 2.2 \\
global threshold & 60.8 $\pm$ 1.6 & 61.3 $\pm$ 2.0 & 50.6 $\pm$ 3.0 \\
\bottomrule
\end{tabular}
\vspace{-12pt}
\end{wraptable}

\textbf{Policy agent.} To isolate the effect of learnability feedback, we remove the Policy Agent from the search loop (Table~\ref{tab:ablation_s1}). Without this signal, the search drives formulation expressiveness higher (Oracle Acc.\ $91.5$), yet this gain does not transfer to test performance, with Total Avg.\ dropping by $3.1$ points. This indicates that expressiveness alone is not a sufficient search signal; policy-side diagnostics help steer the search toward formulations that remain usable after policy instantiation.

\textbf{Oracle agent.} Examining the role of expressiveness feedback, we remove the Oracle Agent from the search loop. Guided solely by policy feedback, the search tends to stay within learnable formulations without exploring sufficiently expressive directions---despite a comparable oracle accuracy level ($87.3$), Total Avg.\ drops by $4.1$ points, a larger drop than removing the Policy Agent. This suggests that the Oracle Agent's structural signal is essential for guiding the search toward expressive formulations.

\begin{wrapfigure}[23]{r}{0.42\textwidth}
  \centering
  \vspace{-12pt}
  \includegraphics[width=0.42\textwidth]{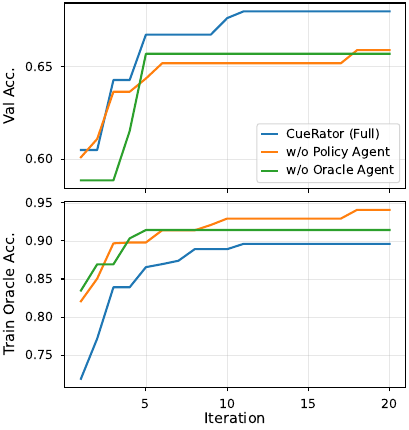}
  \vspace{-19pt}
  \caption{\textbf{Best score trajectory.} Running-best Val Acc.\ and Train Oracle Acc.\ over 20 iterations for the validation-selected session of each variant; Oracle Acc.\ is measured on training data. With oracle--policy feedback, CueRator attains the highest Val Acc.\ despite a lower oracle ceiling.}
  \label{fig:trajectory}
  \vspace{-15pt}
\end{wrapfigure}

\textbf{Score-only loop.} Prior LLM-guided search~\citep{llmsr, funsearch} typically refines candidates with a single LLM guided by a scalar evaluation score. We instantiate this setting by replacing the role-decomposed Plan--Design--Oracle--Policy loop with a single agent that receives only the validation score of each prior candidate, under an otherwise identical protocol. This variant underperforms the full search by $5.8$ points, indicating that search quality stems from the agents' diagnostic reports rather than score-guided refinement alone.

\textbf{No feedback.} Removing both agent feedbacks entirely reduces the search to blind exploration over the symbolic form space, with the Plan Agent seeing only the history of prior formulations. This yields the largest Total Avg.\ drop among search variants at $7.6$ points, demonstrating that the iterative feedback loop is the primary source of search quality, not the agent's intrinsic priors over good formulations. The gap relative to each single-agent removal suggests that both feedback signals each contribute to search quality.

\textbf{Global threshold.} For this variant, we replace the per-video policy with a single global $\boldsymbol{\theta}$ found by oracle search on the training set, with the best candidate selected by validation Acc.\ under this fixed parameter set. Applying the selected formulation to the test data yields a Total Avg.\ of $50.6$, below the globally tuned AV$^2$A rule in Table~\ref{tab:main_result}. The $8.8$-point improvement from the full model then shows that $\pi_\phi$ meaningfully advances performance beyond what global optimization can achieve.

\textbf{Search trajectory.} Fig.~\ref{fig:trajectory} traces the search for the full CueRator and the two single-agent ablations. Without the Policy Agent, late-iteration oracle gains do not transfer to Val Acc., indicating expressive structures that the per-video policy cannot realize; without the Oracle Agent, both metrics plateau early, as the search lacks the structural signal needed to escape learnable but expressively limited regions. The full search rises more slowly in Train Oracle Acc.\ while continuing to improve Val Acc., and by the final iteration it attains the highest Val Acc.\ among the three variants: the dual feedback steers the search toward formulations whose expressiveness is realizable by the policy, rather than toward the highest oracle ceiling.

\needspace{4\baselineskip}
\subsection{Cross-Task Generalization}
\label{sec:generalization}

\begin{table}[t]
\centering
\caption{\textbf{Cross-task generalization.} Baseline scores are as published. For each task, CueRator re-runs the formulation search with the same frozen encoders as the corresponding baseline; search and policy training use seen categories only. Full protocols are in Appendix~\ref{app:gen}.}
\vspace{0.5em}
\label{tab:generalization}
\small
\begin{tabular}{l l l c c c}
\toprule
\multirow{2}{*}[-0.5ex]{Benchmark} & \multirow{2}{*}[-0.5ex]{Baseline} & \multirow{2}{*}[-0.5ex]{Metric} & \multirow{2}{*}[-0.5ex]{Seen:Unseen} & \multicolumn{2}{c}{Performance} \\
\cmidrule(lr){5-6}
 & & & & Baseline & CueRator \\
\midrule
\multirow{2}{*}{OV-DAVEL} & \multirow{2}{*}{Open-DAVTR} & \multirow{2}{*}{mAP} & 75:25 & 22.9 & \textbf{32.0} \\
 & & & 50:50 & 19.4 & \textbf{32.7} \\
\midrule
OV-AVVP & AV$^2$A & Seg.\ Type@AV & 17:8 & 52.4 & \textbf{55.6} \\
\bottomrule
\end{tabular}
\end{table}

We assess generality on two adjacent but structurally different open-vocabulary settings: OV-DAVEL, dense localization of multiple events in untrimmed videos, and OV-AVVP, audio-visual video parsing with weak video-level supervision and modality-wise multi-label outputs. We use the OV-DAVEL benchmark and splits of~\citet{ovdavel} and construct an OV-AVVP benchmark from LLP~\citep{avvp} with a fixed 17-seen/8-unseen split (Appendix~\ref{app:gen}). CueRator improves over Open-DAVTR by $+9.1$ and $+13.3$ mAP on the OV-DAVEL 75:25 and 50:50 splits, and over AV$^2$A by $+3.2$ Seg.\ Type@AV on OV-AVVP (Table~\ref{tab:generalization}). Since each formulation is rediscovered from scratch, these gains indicate that the discovery framework generalizes across video length, supervision regime, and output structure. We also transfer $F^\star$ itself, retraining only the policy. It remains competitive with different encoders on OV-AVEBench, indicating that the rule captures relations among the similarity scores rather than encoder-specific values. Across tasks, it remains competitive on OV-DAVEL but falls short on OV-AVVP, whose output structure differs and which benefits from rediscovery (Appendix~\ref{app:transfer}).

\subsection{Discovered Formulation}
\label{sec:fform}

The selected formulation $F^\star$ computes the audio and visual thresholds as
\begin{align}
\tau_{tc}^{a} &= \theta_0 + \theta_1\,\mathcal{G}(\mathbf{s}^{v})_{tc} + \theta_2\,\mathcal{G}(\mathbf{b}^{a})_{tc} + \theta_3\,\mathcal{G}(\bar{\mathbf{b}}^{a})_{c} + \theta_4\,\beta_{tc}^{a}, \label{eq:tau_a}\\
\tau_{tc}^{v} &= \theta_5 + \theta_6\,\mathcal{G}(\mathbf{s}^{a})_{tc} + \theta_7\,\mathcal{G}(\mathbf{b}^{v})_{tc} + \theta_8\,\mathcal{G}(\bar{\mathbf{b}}^{v})_{c} + \theta_9\,\beta_{tc}^{v}, \label{eq:tau_v}
\end{align}
where $\mathcal{G}(\cdot)$ is the score difference between the strongest competing category and $c$, $\mathbf{b}^{m}$ is the modality score averaged with the cross-modal minimum, $\bar{\mathbf{b}}^{m}$ is its average over the clip, and $\beta^{m}$ is the segment's deviation from that average (exact forms in Appendix~\ref{app:fform}). With the non-negative ranges assigned to $\theta_{1}$--$\theta_{4}$ and $\theta_{6}$--$\theta_{9}$, a category's threshold rises whenever a competitor scores higher, in the other modality, at the segment, or over the clip, or when the segment spikes above its clip average, and falls otherwise: a category is accepted only when both modalities support it against its competitors, with the policy setting how strongly each cue counts per video.
\section{Limitations and Future Directions}
\label{lim}

The current agents reason over textual reports and aggregate statistics rather than inspecting raw audio-visual samples, which may bias the search toward failure modes that appear in these statistics. In addition, the formulation interface targets decisions over cross-modal similarities; richer interactions may require a broader interface. Future work should develop dedicated multimodal agent interfaces.
\section{Conclusion}
\label{con}

We proposed CueRator, an agentic framework for policy-aware decision-rule discovery that adapts frozen contrastive multimodal encoders: a report-guided multi-agent loop searches for a symbolic formulation, and a lightweight policy predicts its parameters per video. On OV-AVEBench, CueRator outperforms training-free, fine-tuning, and dynamic-thresholding baselines with a small seen--unseen gap, and the search is most effective when guided jointly by the expressive ceiling and policy learnability. It also improves over the respective baselines on OV-DAVEL and OV-AVVP, and extending it to other adaptation problems over contrastive multimodal encoders is left to future work.

\clearpage
\appendix
\begin{center}
{\Large\bfseries Appendix}
\vspace{1em}

\textbf{Contents}
\vspace{0.5em}

\renewcommand{\arraystretch}{1.2}
\begin{tabular}{@{}l@{\hspace{1em}}l@{\hspace{1em}\dotfill\hspace{0.5em}}r@{}}
\ref{app:pol} & Adaptive Parameterization Details & \pageref{app:pol} \\
& \quad \ref{app:polarch} Policy Architecture & \pageref{app:polarch} \\
& \quad \ref{app:poll} Policy Learning & \pageref{app:poll} \\[0.3em]
\ref{app:expd} & Additional Experimental Details & \pageref{app:expd} \\[0.3em]
\ref{app:gen} & Cross-Task Generalization Details & \pageref{app:gen} \\[0.3em]
\ref{app:cost} & Computational Cost & \pageref{app:cost} \\[0.3em]
\ref{app:aged} & Agent Details & \pageref{app:aged} \\
& \quad \ref{app:plan_agent} Plan Agent & \pageref{app:plan_agent} \\
& \quad \ref{app:design_agent} Design Agent & \pageref{app:design_agent} \\
& \quad \ref{app:oracle_agent} Oracle Agent & \pageref{app:oracle_agent} \\
& \quad \ref{app:policy_agent} Policy Agent & \pageref{app:policy_agent} \\[0.3em]
\ref{app:fform} & Discovered Formulation & \pageref{app:fform} \\[0.3em]
\ref{app:transfer} & Transferability of the Discovered Formulation & \pageref{app:transfer} \\[0.3em]
\ref{app:sens} & Sensitivity Analyses & \pageref{app:sens} \\[0.3em]
\ref{app:qual} & Qualitative Analysis & \pageref{app:qual} \\[0.3em]
\end{tabular}
\end{center}

\newpage

\section{Adaptive Parameterization Details}
\label{app:pol}

The policy network predicts video-specific parameters for a fixed symbolic formulation. It does not update the pretrained encoders or alter the final audio--visual decision rule. Instead, for each video, it outputs a bounded parameter vector $\boldsymbol{\theta}$ that is consumed by the selected formulation.

\subsection{Policy Architecture}
\label{app:polarch}

\paragraph{Input representation.}
For each video, the policy receives the frozen per-segment audio and visual embeddings
\[
\mathbf{e}^a \in \mathbb{R}^{T \times D},
\qquad
\mathbf{e}^v \in \mathbb{R}^{T \times D},
\]
as defined in Section~\ref{sec:problem-setup}. The policy does not take category text embeddings as input; text embeddings are used by the symbolic formulation and the final threshold-based decision rule. This keeps the policy architecture independent of the category vocabulary and allows the same policy design to be used across different open-vocabulary category sets.

\paragraph{Architecture.}
The audio and visual embeddings are first projected into a shared hidden dimension $d_h$ using modality-specific linear layers:
\[
\tilde{\mathbf{e}}^a = \mathrm{Linear}_a(\mathbf{e}^a),
\qquad
\tilde{\mathbf{e}}^v = \mathrm{Linear}_v(\mathbf{e}^v).
\]
We use $d_h=256$ in all experiments.

Each modality is then processed by a modality-specific Transformer-style self-attention block:
\[
\mathbf{h}^a = \mathrm{SelfAttn}_a(\tilde{\mathbf{e}}^a),
\qquad
\mathbf{h}^v = \mathrm{SelfAttn}_v(\tilde{\mathbf{e}}^v).
\]
We use one self-attention block per modality. Each block consists of multi-head attention, residual connection, layer normalization, and a feed-forward network with GELU activation. All attention blocks use $4$ heads, a $512$-dimensional feed-forward layer, and dropout $0.1$.

To exchange information between modalities, we apply bidirectional cross-attention:
\[
\hat{\mathbf{h}}^v
=
\mathrm{CrossAttn}_v(\mathbf{h}^v, \mathbf{h}^a),
\qquad
\hat{\mathbf{h}}^a
=
\mathrm{CrossAttn}_a(\mathbf{h}^a, \mathbf{h}^v),
\]
where the first argument provides the query sequence and the second argument provides keys and values. Thus, the visual stream attends to the audio stream, and the audio stream attends to the visual stream through separate cross-attention blocks.

The resulting temporal features are mean-pooled over segments:
\[
\bar{\mathbf{h}}^a
=
\frac{1}{T}\sum_{t=1}^{T}\hat{\mathbf{h}}^a_t,
\qquad
\bar{\mathbf{h}}^v
=
\frac{1}{T}\sum_{t=1}^{T}\hat{\mathbf{h}}^v_t.
\]
The pooled modality features are concatenated and fused:
\[
\mathbf{h}
=
\mathrm{Dropout}
\left(
\mathrm{GELU}
\left(
W_f[\bar{\mathbf{h}}^a;\bar{\mathbf{h}}^v] + \mathbf{b}_f
\right)
\right).
\]
A linear head predicts the mean of a Gaussian action distribution,
\[
\boldsymbol{\mu} = W_\mu \mathbf{h} + \mathbf{b}_\mu \in \mathbb{R}^{P},
\]
where $P$ is the number of parameters defined by the selected formulation.

\paragraph{Range-bounded parameterization.}
The policy operates in an unconstrained raw action space. For each parameter dimension $i$, the selected formulation specifies a valid range $[\theta_i^{\min}, \theta_i^{\max}]$. A raw action $z_i$ is mapped into this range by
\[
\theta_i
=
\theta_i^{\min}
+
\sigma(z_i)
\left(
\theta_i^{\max}-\theta_i^{\min}
\right).
\]
This guarantees that the predicted parameters remain inside the formulation-defined bounds.

During training, raw actions are sampled independently for each parameter dimension from a Gaussian centered at $\boldsymbol{\mu}$, as described in Appendix~\ref{app:poll}. During evaluation, no sampling is used; the deterministic mean action $\boldsymbol{\mu}$ is mapped directly to $\boldsymbol{\theta}$ and passed to the selected formulation.

\subsection{Policy Learning}
\label{app:poll}

We train the policy as a contextual bandit. For a video $n$, the context is the frozen audio--visual embedding sequence $(\mathbf{e}^a_n,\mathbf{e}^v_n)$, the action is the formulation parameter vector $\boldsymbol{\theta}$, and the reward is the task score obtained after applying the fixed formulation and the final audio--visual decision rule:
\[
r_{n,k}
=
\mathcal{R}(F,\boldsymbol{\theta}_{n,k};n).
\]
In our implementation, $\mathcal{R}$ is per-video segment accuracy, chosen because it is fast to compute for each sampled action in the policy-gradient inner loop. Final results are reported using the full OV-AVEL metric suite.

\paragraph{Action sampling.}
The policy predicts the mean $\boldsymbol{\mu}_n \in \mathbb{R}^{P}$ in the raw action space. During training, we sample $K$ raw action vectors $\mathbf{z}_{n,k}\in\mathbb{R}^{P}$. Each parameter dimension is sampled independently as
\[
z_{n,k,i}
\sim
\mathcal{N}(\mu_{n,i},\sigma_a^2),
\qquad
k=1,\ldots,K,\quad i=1,\ldots,P,
\]
with fixed $\sigma_a=0.3$. Each raw action vector is mapped to the formulation-defined range by the sigmoid transformation in Appendix~\ref{app:polarch}, producing $\boldsymbol{\theta}_{n,k}$.

\paragraph{Leave-one-out REINFORCE.}
Because the thresholding and agreement rule is non-differentiable, gradients do not flow through the reward. We therefore optimize the policy with REINFORCE. For each sampled raw action vector, we use the average reward of the other samples from the same video as a leave-one-out baseline:
\[
b_{n,k}
=
\frac{1}{K-1}
\sum_{j\neq k}
r_{n,j}.
\]
The resulting advantage is
\[
\hat{A}_{n,k}
=
r_{n,k}-b_{n,k}.
\]
The policy-gradient loss over a minibatch $\mathcal{B}$ is
\[
\mathcal{L}_{\mathrm{PG}}
=
-\frac{1}{|\mathcal{B}|K}
\sum_{n\in\mathcal{B}}
\sum_{k=1}^{K}
\hat{A}_{n,k}
\sum_{i=1}^{P}
\log
\mathcal{N}(z_{n,k,i};\mu_{n,i},\sigma_a^2).
\]
In implementation, the advantages are treated as constants, so gradients are taken only through the Gaussian log-probability terms.
This objective directly optimizes the task reward induced by the same hard decision rule used at evaluation time, without requiring gradients through the thresholding and agreement rule.

\paragraph{Training details.}
We use $K=4$ samples per video, AdamW with learning rate $1\times10^{-4}$, batch size $32$, and train for $20$ epochs. For each candidate formulation, the checkpoint with the highest validation accuracy is selected.

\section{Additional Experimental Details}
\label{app:expd}

This section describes how we produced the AV$^2$A baselines in Table~\ref{tab:main_result}. The dataset split, evaluation metrics, encoder choice, and CueRator training protocol are given in the main paper and Appendix~\ref{app:pol}; here we specify only the AV$^2$A adaptation and hyperparameter selection procedure.

\paragraph{Why we re-evaluate AV$^2$A.}
We re-evaluate AV$^2$A for two reasons. First, the original AV$^2$A paper does not report results on OV-AVEBench, so there is no directly comparable number for the benchmark used in Table~\ref{tab:main_result}. Second, our main experiments use ImageBind, whereas the original paper reports CLIP+CLAP and LanguageBind only. We tune the AV$^2$A global hyperparameters on a validation subset, following the protocol of the original paper.

\paragraph{Hyperparameter selection.}
The original AV$^2$A paper selects its hyperparameters on 200 validation videos. We follow this protocol on OV-AVEBench: from the 5{,}798 validation videos we draw a class-stratified subset of 200 videos (seed 42) that covers all 67 categories with at least one video each (61 videos of seen categories and 139 of unseen categories). We evaluate 8{,}192 hyperparameter combinations generated by a Sobol sequence (seed 42) over $\alpha,\tau^0,\tau_r,\tau_f \in [0,1]$ and $\lambda \in [0,5]$, and select the combination with the highest mean of Acc., Seg., and Eve.\ on this subset. The selected combination is $\alpha=0.52$, $\tau^0=0.96$, $\tau_r=0.37$, $\tau_f=0.18$, and $\lambda=4.55$. The selected values are kept fixed for all videos and categories when reporting test metrics; the test split is not used for selection.

\paragraph{AV$^2$A-Policy.}
AV$^2$A-Policy is a controlled variant that isolates the effect of per-video parameter prediction from that of the discovered formulation. It retains AV$^2$A's manually designed decision rule, but replaces the globally fixed hyperparameters $(\alpha,\tau^0,\tau_r,\tau_f,\lambda)$ with per-video predictions from a policy network. The policy uses the same backbone architecture, frozen ImageBind embeddings, leave-one-out REINFORCE training, hyperparameters, and validation-based checkpoint selection as CueRator (Appendix~\ref{app:pol}); only the parameter vector it outputs is redefined to instantiate the AV$^2$A rule. All five parameters are produced through the same sigmoid mapping as in Appendix~\ref{app:polarch}, with $(\alpha,\tau^0,\tau_r,\tau_f)$ bounded to $[0,1]$ and $\lambda$ to $[0,5]$. Training uses seen-category videos only, matching the CueRator protocol.

\section{Cross-Task Generalization Details}
\label{app:gen}

This section provides the full protocols for the cross-task generalization experiments in Section~\ref{sec:generalization}. As in the main experiments, each task runs three isolated 20-iteration search sessions, and the candidate with the best validation performance across sessions is reported; the test split remains held out for final evaluation. Transfer of the fixed OV-AVEBench formulation to these tasks is analyzed separately in Appendix~\ref{app:transfer}.

\paragraph{OV-DAVEL.}
OV-DAVEL~\citep{ovdavel} extends dense audio-visual event localization~\citep{davel}, which localizes multiple, possibly overlapping events in untrimmed videos, to the open-vocabulary setting, in which the test vocabulary includes categories unseen during training. We follow the official 75:25 and 50:50 seen:unseen category splits and report mAP, comparing against the one-stage Open-DAVTR baseline~\citep{ovdavel}. We re-run the full CueRator formulation search using frozen ViT-Lens~\citep{vitlens} encoders, matching Open-DAVTR, with each segment spanning 0.32\,s.

\paragraph{Adapting to event-level evaluation.}
Since mAP requires scored event proposals while the OV-AVEL decision rule yields only hard per-segment labels, we form events by merging temporally contiguous segments assigned the same category, and extend the search loop so that each candidate jointly specifies a confidence formulation that maps the audio-text and visual-text similarities within an event span to an event-level confidence. The parameters of this confidence formulation are not predicted per video; they are fitted once on the training split by Sobol search. Formulation discovery, policy training, and confidence-parameter fitting use seen categories only.

\paragraph{OV-AVVP.}Audio-visual video parsing (AVVP)~\citep{avvp} predicts modality-wise, multi-label events under weak video-level supervision, a supervision regime and output structure that differ substantially from OV-AVEL. Because no standard open-vocabulary split exists for AVVP, we construct one from the 25 LLP categories~\citep{avvp}: a fixed 17-seen/8-unseen split, randomly generated once before any experiment. We use frozen LanguageBind~\citep{languagebind} encoders, following the best published AV$^2$A configuration. All methods are evaluated on the official LLP test videos with the standard AVVP metrics; we report segment-level Type@AV. AV$^2$A~\citep{av2a} is training-free, so the seen/unseen split does not affect its score; its published result on the official LLP test videos is therefore directly comparable.

\paragraph{Adapting to weak supervision.}Modality-wise predictions follow directly from the per-modality thresholds of the decision rule (Section~\ref{sec:problem-setup}); since AVVP allows multiple concurrent events, we output every category that passes a modality's threshold rather than only the top-scoring one. The main difficulty is weak supervision: LLP provides only video-level event labels for training, so the per-video reward used for policy training (Appendix~\ref{app:poll}) is unavailable. We therefore extend the search loop with two components. First, each candidate jointly specifies a pseudo-label formulation that maps the audio-text and visual-text similarities, together with the video-level label, to per-segment pseudo-labels that serve as the policy's training targets; its parameters are fitted globally by full-batch gradient descent on the official validation split, which, unlike the training split, has segment-level labels. Second, we reuse the event-level confidence formulation of OV-DAVEL as a filter: a drop threshold on the confidence is selected by grid search on the validation split, and event spans below it are removed, both from the pseudo-labels and from the final predictions. AV$^2$A likewise tunes its hyperparameters on validation videos; the test split is used by neither method. Formulation discovery, policy training, and all parameter fitting use seen categories only.

\section{Computational Cost}
\label{app:cost}

Table~\ref{tab:wallclock} reports the per-session wall-clock breakdown of the agentic search, averaged over the 3 search sessions in our main experiments. Each session runs $20$ iterations on a single RTX 3090, with LLM calls (Plan, Design, Oracle, Policy agents) issued through gpt-5.4 in OpenAI Codex (v0.125.0, April 2026) at high reasoning effort.

\begin{table}[h]
\centering
\caption{Per-session wall-clock breakdown of the agentic search ($20$ iterations), averaged over $3$ sessions on a single RTX 3090.}
\label{tab:wallclock}
\small
\begin{tabular}{lcc}
\toprule
Component & Time (h) & Share \\
\midrule
Agent (LLM inference)         & 4.0 &  41\% \\
Oracle search                 & 1.8 &  18\% \\
Policy training \& evaluation & 4.0 &  41\% \\
\midrule
Total                         & 9.8 & 100\% \\
\bottomrule
\end{tabular}
\end{table}

The oracle search remains efficient despite using a Sobol sequence of $2^{17}$ samples per video, as ImageBind embeddings are precomputed and cached, and the per-sample formulation evaluation is parallelized on GPU.

\section{Agent Details}
\label{app:aged}

This section documents the agentic search loop used to discover the
symbolic formulation. The goal is not to prescribe a particular natural
language style, but to make clear what information each agent receives,
what computation it is allowed to perform, and what artifact is returned
to the shared memory. We therefore describe each agent through three
parts: \emph{prompt input}, \emph{operation}, and \emph{output}. Table~\ref{tab:gloss} summarizes the terminology used by the search loop.

\begin{table}[h]
\centering
\caption{Glossary of CueRator search terminology.}
\label{tab:gloss}
\small
\renewcommand{\arraystretch}{1.2}
\begin{tabularx}{\textwidth}{@{}p{0.24\textwidth} X@{}}
\toprule
Term & Meaning \\
\midrule
Formulation $F$ & A closed-form rule computing per-segment thresholds from the audio, visual, and text embeddings (via their similarities), with $P$ tunable parameters $\boldsymbol{\theta}$. \\
Oracle ceiling & The performance of $F$ when $\boldsymbol{\theta}$ is set optimally for each video---an upper bound on what the rule's structure can express. \\
Policy network $\pi_\phi$ & A small network predicting $\boldsymbol{\theta}$ from the audio and visual embeddings, called a \emph{policy} because it is trained with REINFORCE (thresholding is non-differentiable). \\
Learnability & Whether the optimal $\boldsymbol{\theta}$ is predictable from the video---measured as the gap between the oracle ceiling and the policy's performance. \\
Reports / Memory / Directives & The agents' written analyses (\emph{reports}) accumulate across iterations (\emph{memory}); the Plan Agent reads them and issues instructions for the next iteration (\emph{directives}). \\
\bottomrule
\end{tabularx}
\end{table}

\begin{table*}[t]
\centering
\small
\setlength{\tabcolsep}{4pt}
\renewcommand{\arraystretch}{1.15}
\caption{
Input--operation--output view of one CueRator search iteration.
The LLM agents are used only during offline formulation discovery; test-time prediction uses the selected formulation and policy network.
}
\label{tab:agent_roles}
\begin{tabularx}{\textwidth}{@{}p{0.12\textwidth} X X X@{}}
\toprule
Agent & Receives & Performs & Writes to memory \\
\midrule
\textbf{Plan}
& Prior candidate summaries, per-candidate records (formulations, parameter ranges, oracle scores, policy validation scores, agent reports).
& Diagnoses the search state; decides whether to refine, abandon, or redirect formulation families; assigns parameter roles, range changes, ablation targets, and analysis requests.
& Design directive, oracle ablation protocol, policy analysis protocol, and one-line planning summary. \\

\textbf{Design}
& Plan directive, formulation interface, symbolic-form constraints, and prior formulation code patterns.
& Proposes an executable formulation \(F_i(\cdot;\theta)\); declares parameter roles and ranges \(\Theta_i\); revises the code until AST checks pass.
& Candidate code, parameter ranges, parameter role descriptions, validity status, and design summary. \\

\textbf{Oracle}
& Verified formulation \(F_i\), parameter ranges \(\Theta_i\), training split, and requested ablations.
& Searches per-video oracle parameters; estimates expressive capacity; repeats per-video oracle search for each ablation; analyzes parameter distributions, boundary saturation, and component impact.
& Oracle score \(\mathcal{O}(F_i)\), ablation statistics, parameter diagnostics, raw-analysis artifacts, and oracle report. \\

\textbf{Policy}
& Verified formulation \(F_i\), parameter ranges \(\Theta_i\), trained policy outputs, validation metrics, and requested diagnostics.
& Analyzes the policy \(\pi_{\phi_i}\) trained for the candidate: policy-instantiated behavior, predicted parameter distributions, and failure modes.
& Validation score \(\mathcal{V}(F_i,\pi_{\phi_i})\), predicted-parameter statistics, learnability diagnostics, and policy report. \\
\bottomrule
\end{tabularx}
\end{table*}

\paragraph{Reporting convention.}
The examples below are representative excerpts from the actual search
logs. Long file paths, repeated environment boilerplate, and non-essential
token-level traces are omitted for readability. When a log contains
private chain-of-thought or provider-specific hidden reasoning, we report
only the agent-visible plan, executed commands, code snippets, and final
analysis text. This makes the agent-visible inputs and outputs auditable
without relying on unrecoverable internal reasoning.

\subsection{Plan Agent}
\label{app:plan_agent}

\paragraph{Prompt input.}
At the beginning of each search iteration, the Plan Agent receives the
current iteration index and a compact memory of previous iterations. The
memory contains the candidate identifier, validation and oracle metrics
when available, parameter ranges, one-line reports from the Design,
Oracle, and Policy Agents, and pointers to detailed reports. The prompt
also states the fixed formulation interface and asks the Plan Agent to
produce a directive for the next candidate rather than a formulation
itself.

\paragraph{Operation.}
The Plan Agent does not execute model training or modify the formulation
code directly. Instead, it reads the accumulated memory and identifies
which aspects of the search should be changed in the next iteration. In
particular, it can (i) retain structural motifs that have repeatedly
improved oracle or validation performance, (ii) request that ineffective
parameters be removed, repurposed, or range-adjusted, (iii) specify
which parameters should be ablated by the oracle evaluation script, and (iv) ask the
Oracle or Policy Agent to inspect specific failure modes in the next
analysis. This makes the Plan Agent the controller of the search
direction, while leaving executable formulation construction and
empirical analysis to the specialized agents.

\paragraph{Output.}
The Plan Agent outputs a short directive consumed by the Design Agent and
the later analysis agents. The directive typically includes the intended
role of each parameter, the desired parameter ranges or range changes,
the requested oracle ablations, and analysis questions to answer after
the candidate is evaluated.

\begin{figure}
    \centering
    \includegraphics[width=\linewidth,height=0.9\textheight,keepaspectratio]{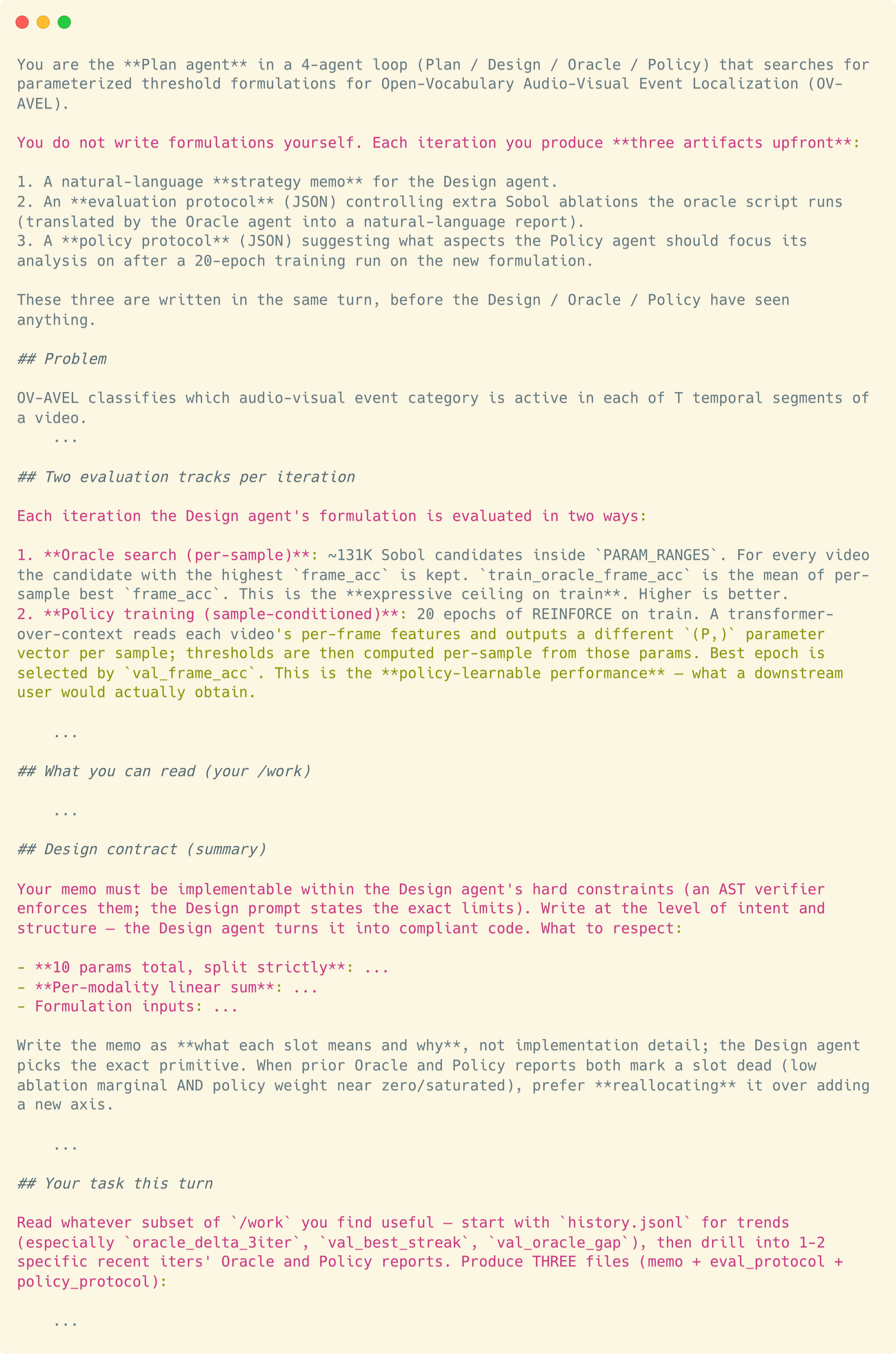}
    \caption{Condensed input prompt for the Plan Agent.}
    \label{fig:planprompt}
\end{figure}

\begin{figure}
    \centering
    \includegraphics[width=1\linewidth]{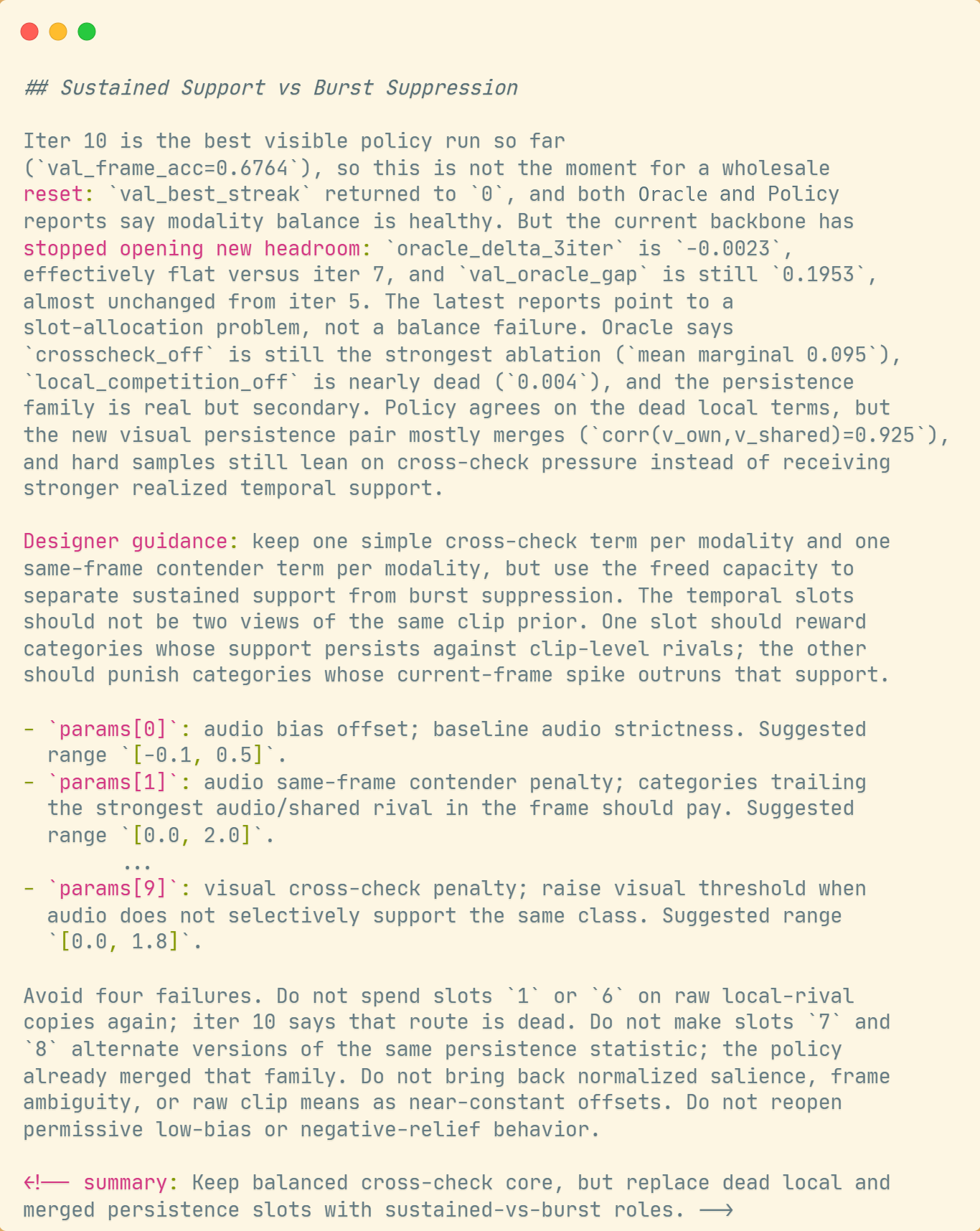}
    \caption{Condensed Strategy directive produced by the Plan Agent for the Design Agent.}
    \label{fig:plandirec}
\end{figure}

\begin{figure}
    \centering
    \includegraphics[width=1\linewidth]{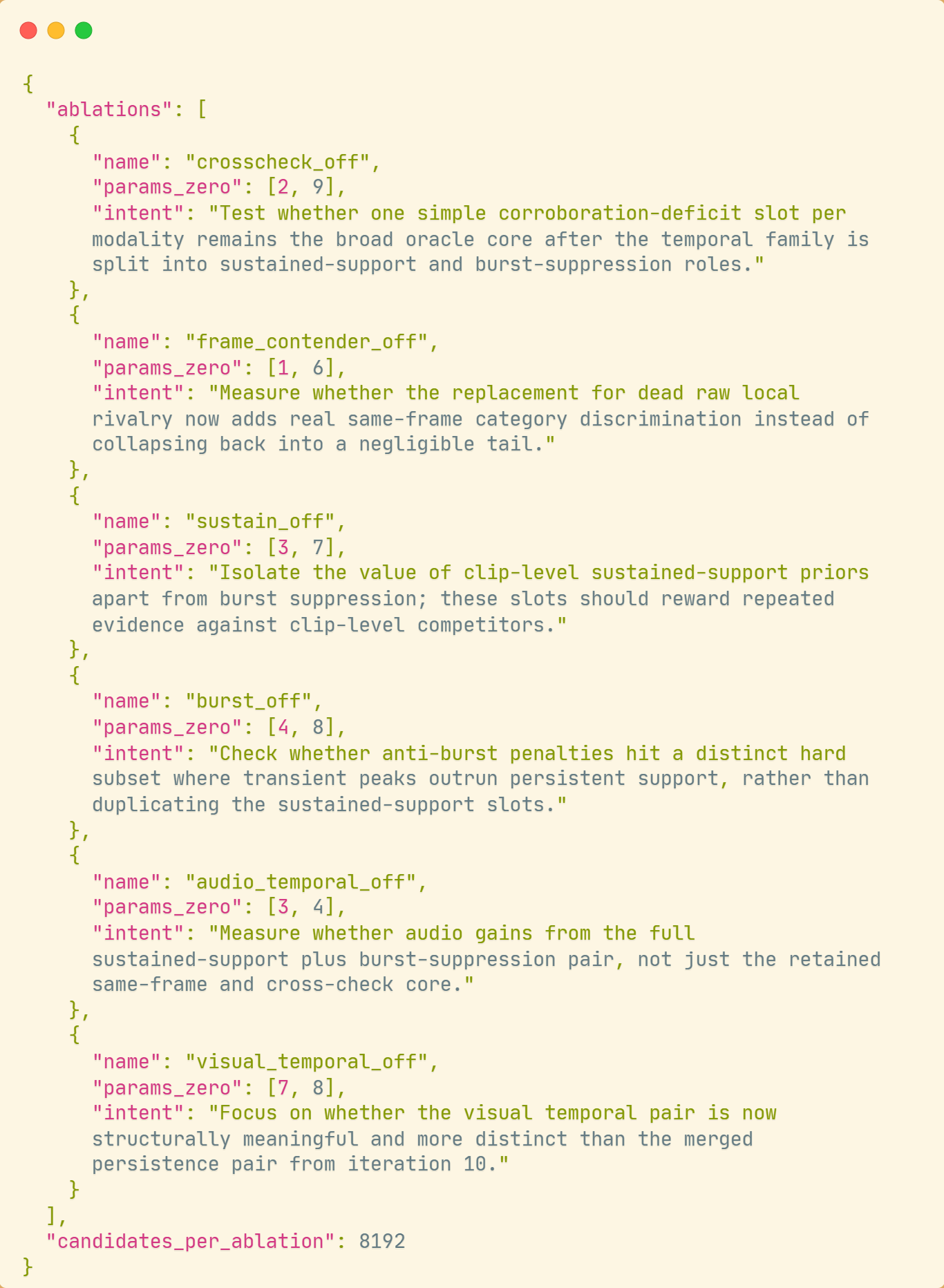}
    \caption{Oracle ablation protocol produced by the Plan Agent.}
    \label{fig:planablation}
\end{figure}

\begin{figure}
    \centering
    \includegraphics[width=1\linewidth]{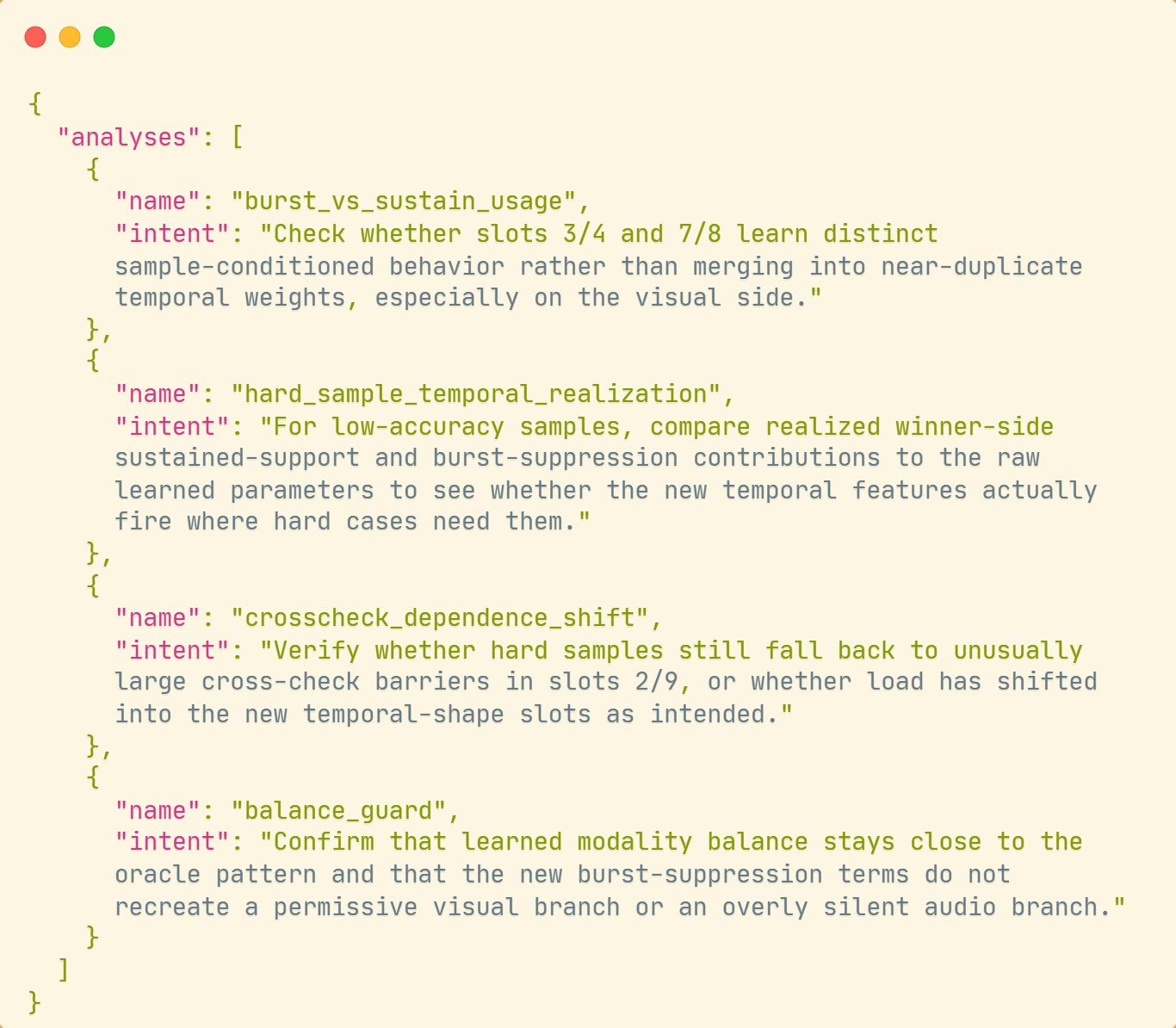}
    \caption{Policy analysis protocol produced by the Plan Agent.}
    \label{fig:plananalysis}
\end{figure}

\newpage

\subsection{Design Agent}
\label{app:design_agent}

\paragraph{Prompt input.}
The Design Agent receives the Plan Agent's strategy memo, the formal
formulation interface, previous formulation files as code-pattern
references, and the symbolic-form constraints. It does not receive the
full search history or oracle/policy reports directly; those signals are
summarized by the Plan Agent. The interface specifies the available
inputs---frozen audio embeddings, visual embeddings, audio text
embeddings, and visual text embeddings---and requires the function to
return audio and visual threshold matrices. The prompt also states the
final decoding rule: categories are filtered by modality-specific
thresholds, the top surviving category is selected in each modality, and
the segment is assigned a foreground category only when the two
modalities agree.

\paragraph{Operation.}
The Design Agent writes an executable Python implementation of the
candidate formulation together with a bounded range for every parameter.
Each candidate is checked by an AST-based validator before evaluation.
The validator enforces the checkable constraints of Section~\ref{sec:design_agent}: no control flow, no parameter used as an inner argument to a term, numeric literals restricted to the allowed set, and the function-count and statement budgets; the linear-form template itself is specified in the prompt. If the validator rejects the candidate, the diagnostic
message is returned to the Design Agent, which revises the function
within the same iteration until a valid candidate is produced or the
iteration fails.

\paragraph{Output.}
The Design Agent outputs a Python formulation file, a parameter range
table, parameter names, and a short natural-language rationale for the
candidate. The rationale records why each term was introduced and which
Plan Agent request it is intended to address. The submitted code, rather
than the rationale, is the object checked by the AST validator and
evaluated by the Oracle and Policy stages.

\begin{figure}
    \centering
    \includegraphics[width=\linewidth,height=0.9\textheight,keepaspectratio]{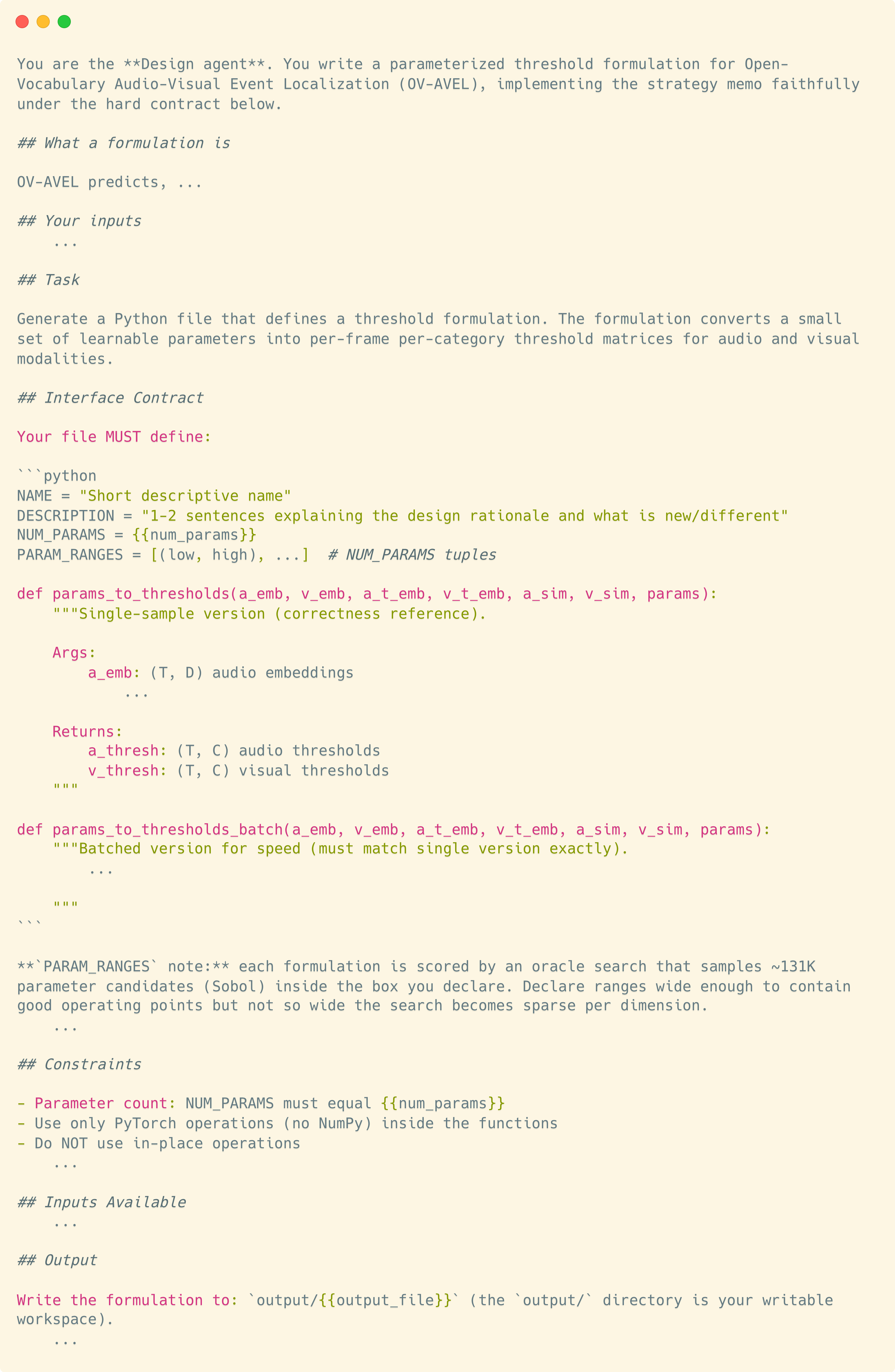}
    \caption{Condensed input prompt for the Design Agent.}
    \label{fig:designprompt}
\end{figure}

\begin{figure}
    \centering
    \includegraphics[width=0.8\linewidth]{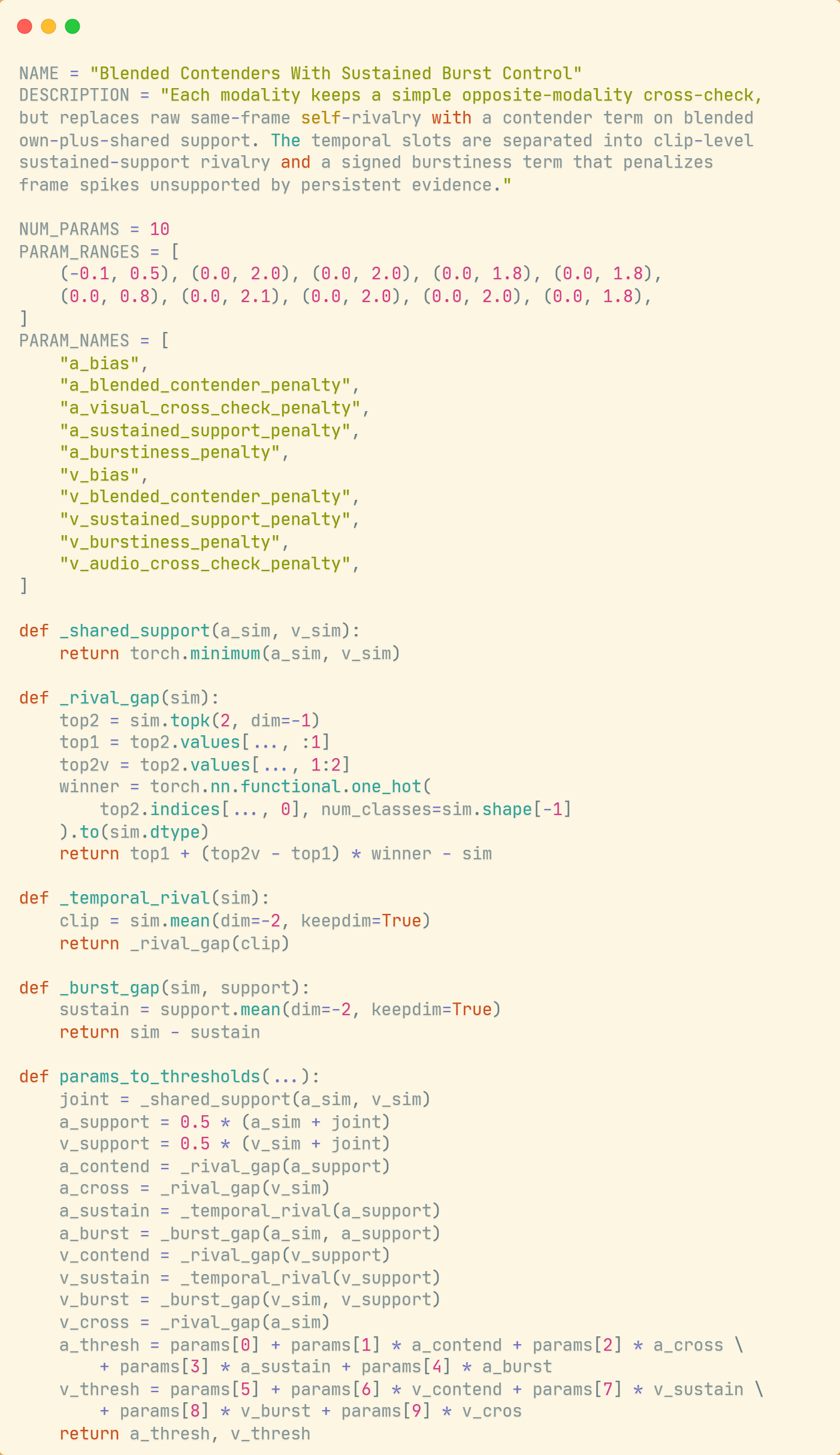}
    \caption{Condensed formulation code produced by the Design Agent.}
    \label{fig:designform}
\end{figure}

\newpage

\subsection{Oracle Agent}
\label{app:oracle_agent}

\paragraph{Prompt input.}
The Oracle Agent receives the verified formulation, its parameter ranges,
the Plan Agent's requested ablations, and the raw outputs of the oracle
search. For every training video, the oracle search records the best
parameter vector found by Sobol sampling and the corresponding per-video
reward. For requested ablations, it also records the performance obtained
after zeroing or disabling designated parameters and re-optimizing the
remaining parameters.

\paragraph{Operation.}
The Oracle Agent analyzes the expressive capacity of the formulation
under ideal per-video parameters. It is allowed to write and execute
custom analysis scripts over the oracle search outputs, for example to
compute parameter distribution statistics, identify boundary-pinned
parameters, measure per-parameter ablation effects, inspect correlations
between parameters and video-level performance, and compare successful
and failed videos. These scripts are not part of the final model; they
serve only to produce diagnostic evidence for the search memory.

\paragraph{Output.}
The Oracle Agent returns a report summarizing what the oracle results
imply about the formulation's structure. The report distinguishes
between high-level performance, parameter-level findings, and diagnostic
signals used by the next Plan Agent. A one-line summary is appended to
the compact memory, while the full report remains available for later
inspection.

\begin{figure}
    \centering
    \includegraphics[width=\linewidth,height=0.9\textheight,keepaspectratio]{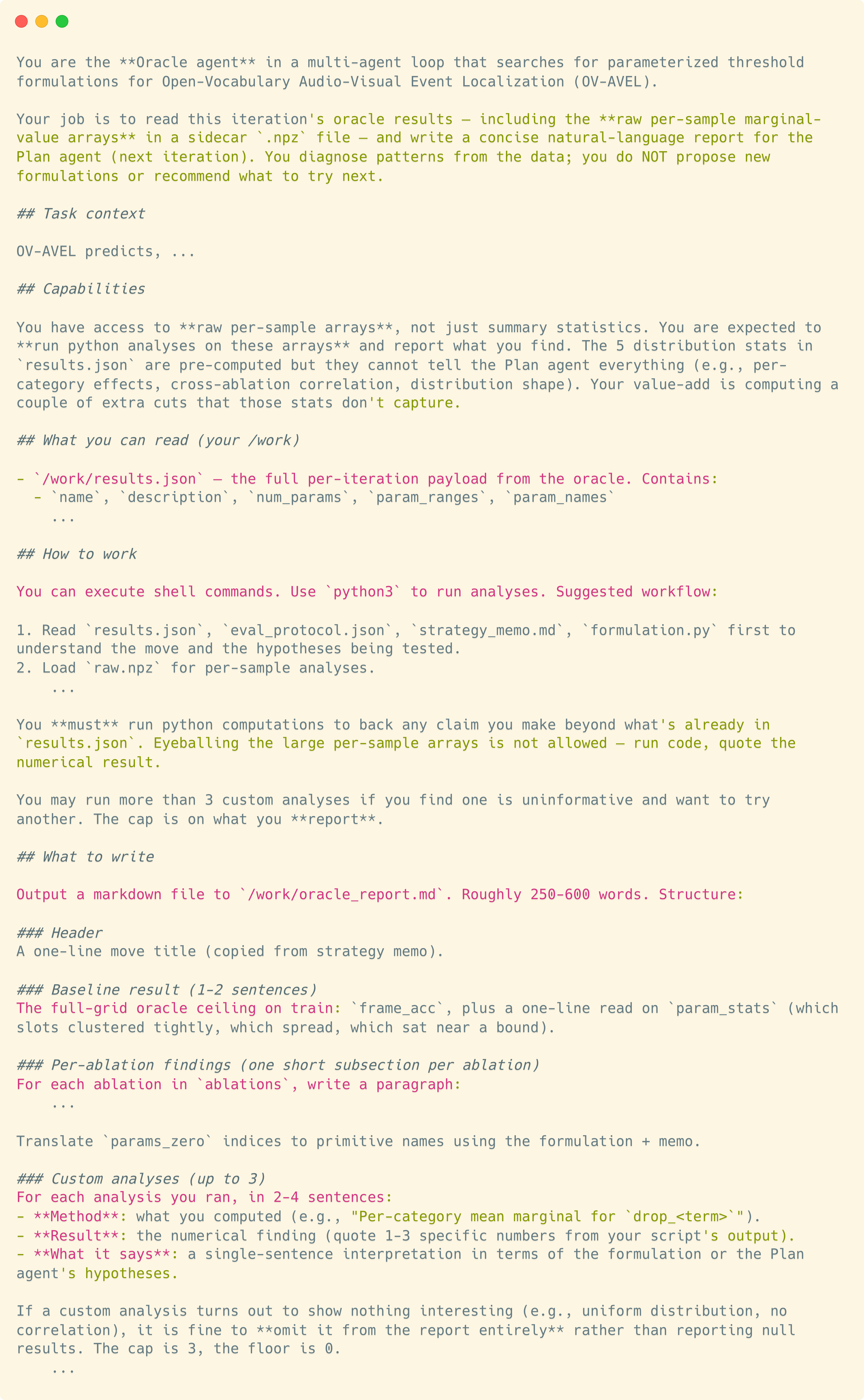}
    \caption{Condensed input prompt for the Oracle Agent.}
    \label{fig:oracleprompt}
\end{figure}

\begin{figure}
    \centering
    \includegraphics[width=1\linewidth]{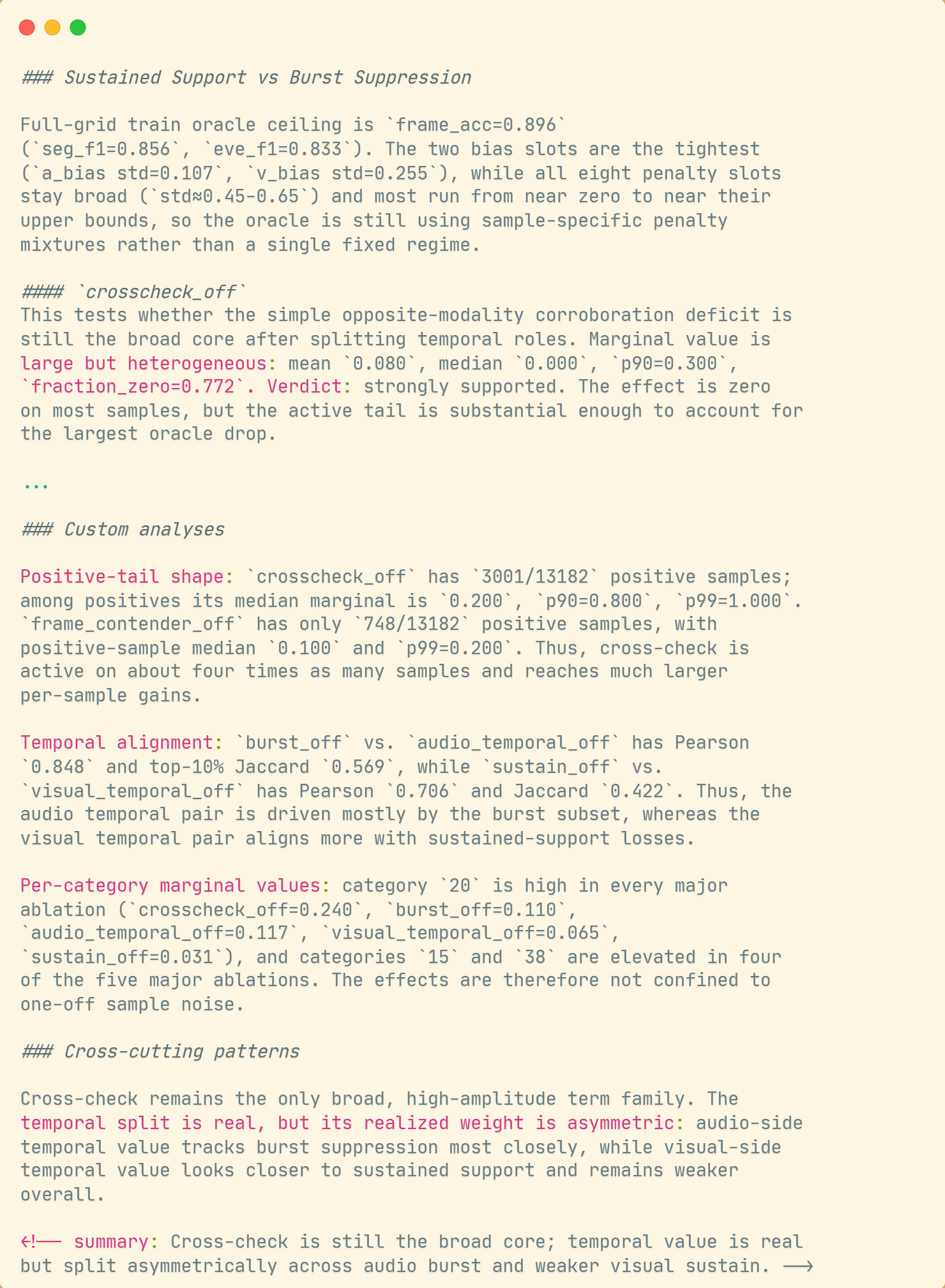}
    \caption{Condensed output report of the Oracle Agent.}
    \label{fig:oraclereport}
\end{figure}

\newpage

\subsection{Policy Agent}
\label{app:policy_agent}

\paragraph{Prompt input.}
The Policy Agent receives the verified formulation, its parameter ranges,
the trained policy checkpoint selected by validation accuracy, and the
stored per-video outputs produced by that checkpoint. These outputs
include predicted parameter vectors, segment-level predictions,
per-video rewards, and aggregate validation metrics. When oracle results
are available for the same candidate, the prompt may also include
pointers to the oracle parameter statistics so that the Policy Agent can
compare what the policy predicts against what the oracle prefers.

\paragraph{Operation.}
The Policy Agent analyzes whether the formulation is learnable by the
policy network, rather than merely expressive under oracle parameters.
It may write and execute scripts to inspect the distribution of predicted
parameters, detect collapsed or boundary-saturated policy outputs,
measure the gap between oracle and policy performance, identify videos
where the policy improves or fails, and summarize modality-specific
prediction patterns. The resulting diagnostics are used to determine
whether a high-oracle formulation is practically useful after policy
training.

\paragraph{Output.}
The Policy Agent returns a report focused on policy learnability. The
report states whether the predicted parameters use the available ranges,
whether the learned policy collapses to a narrow region of the parameter
space, which learnability failure modes dominate, and which diagnostic
signals should be passed to the next planning step. As with the Oracle
Agent, the full report is stored for inspection and a one-line summary is
appended to the compact memory.

\begin{figure}
    \centering
    \includegraphics[width=\linewidth,height=0.9\textheight,keepaspectratio]{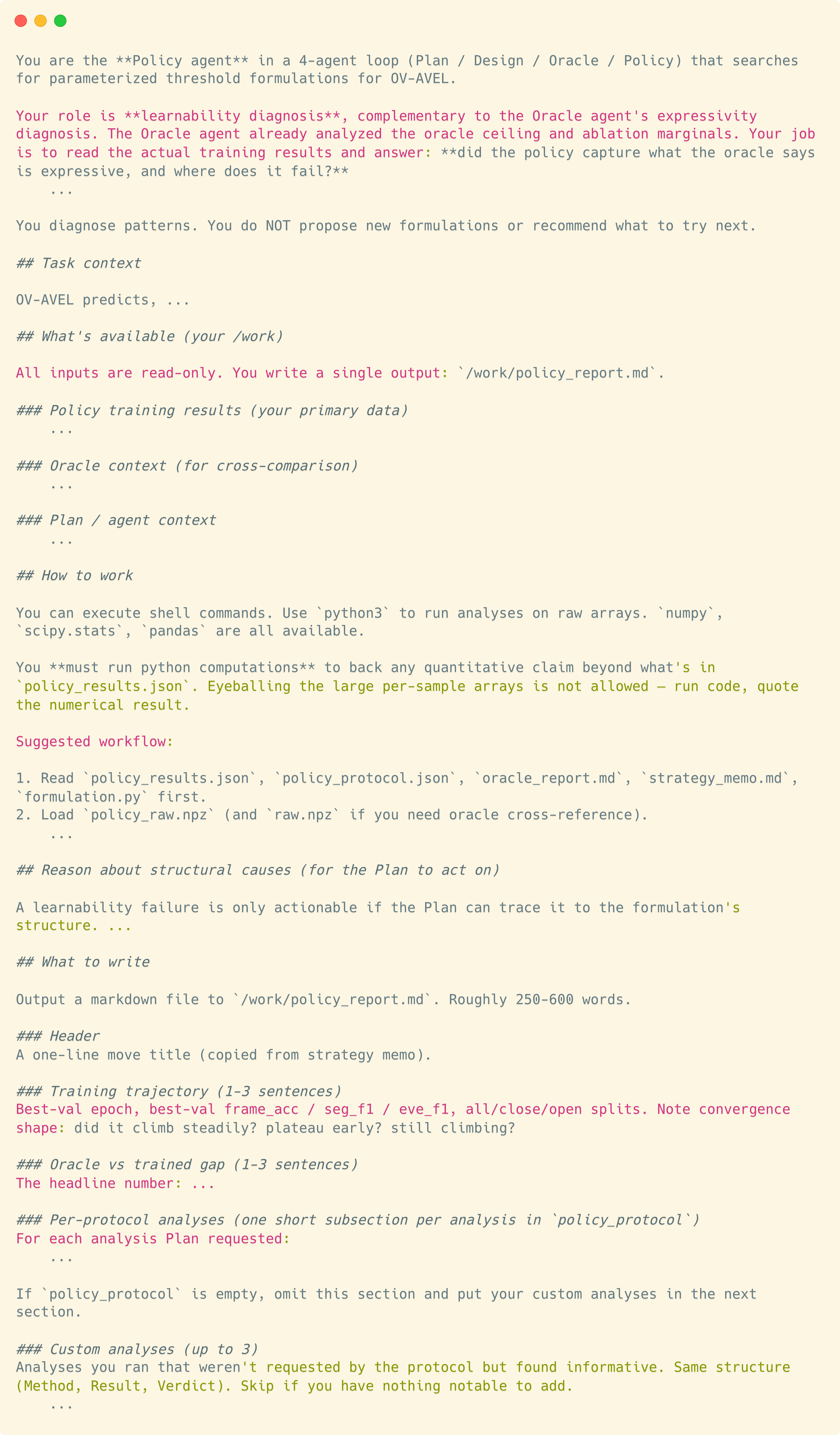}
    \caption{Condensed input prompt for the Policy Agent.}
    \label{fig:policyprompt}
\end{figure}

\begin{figure}
    \vspace{20pt}
    \centering
    \includegraphics[width=1\linewidth]{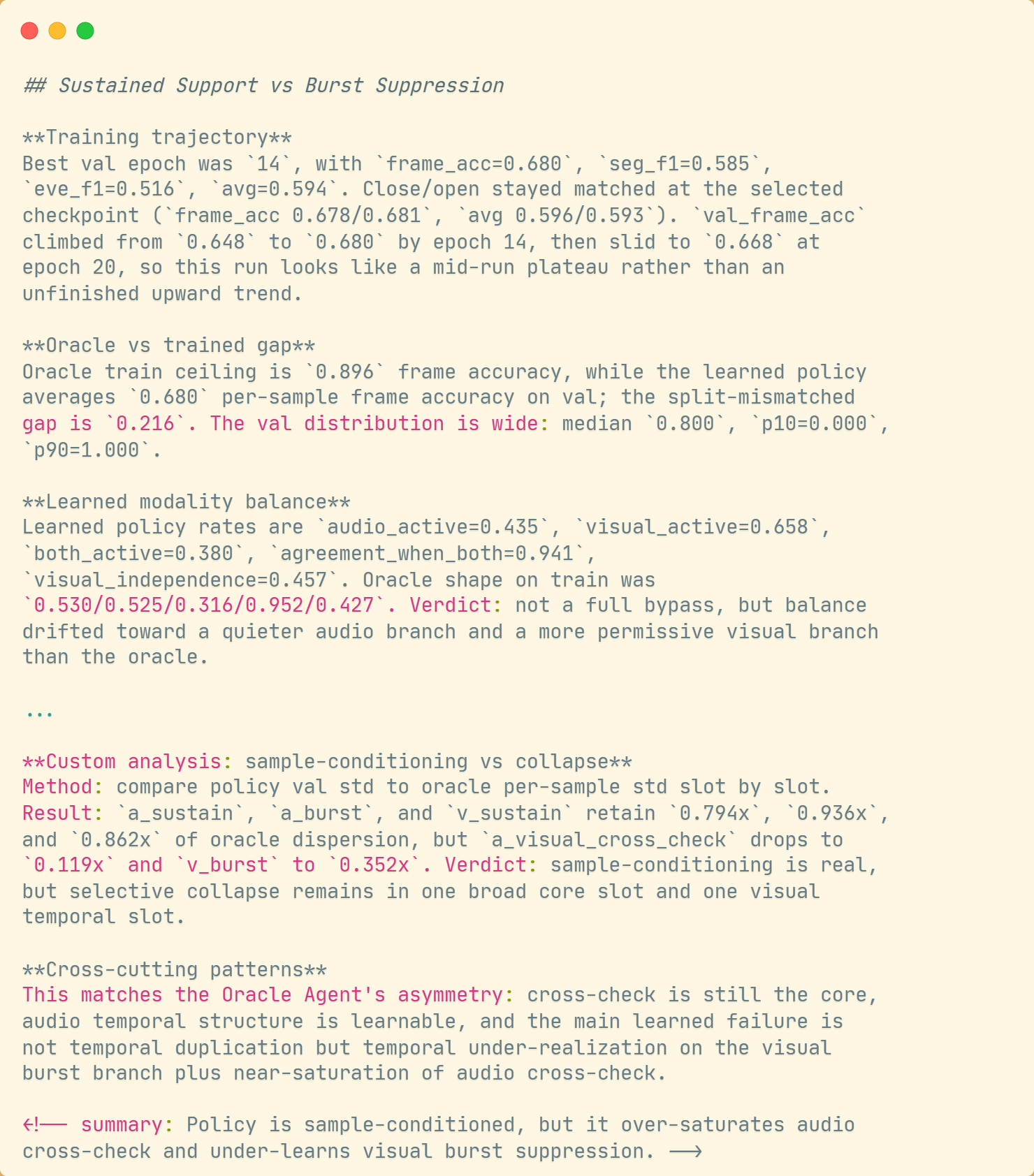}
    \caption{Condensed output report of the Policy Agent.}
    \label{fig:policyreport}
    \vspace{40pt}
\end{figure}

\newpage

\section{Discovered Formulation}
\label{app:fform}

This section gives the exact form of the formulation summarized in Section~\ref{sec:fform}, together with its parameter ranges and an interpretation of each term. We use the notation from Section~\ref{sec:problem-setup}:
$s_{tc}^{m}$ denotes the cosine similarity between modality
$m \in \{a,v\}$ at segment $t$ and category text embedding $c$.
The formulation produces thresholds $\boldsymbol{\tau}^a$ and
$\boldsymbol{\tau}^v$; the subsequent filtering, top-category selection,
and audio--visual agreement rule are exactly the fixed decision procedure
defined in Section~\ref{sec:problem-setup}.

The formulation first computes, for each segment-category pair, the lower
of the audio-text and visual-text similarity scores:
\begin{align*}
q_{tc} &= \min(s_{tc}^{a}, s_{tc}^{v}),
\end{align*}
and averages this value with each modality's original similarity score:
\begin{align*}
b_{tc}^{a} &= \frac{1}{2}(s_{tc}^{a} + q_{tc}), \\
b_{tc}^{v} &= \frac{1}{2}(s_{tc}^{v} + q_{tc}).
\end{align*}
Here $q_{tc}$ is high only when both modalities assign a high similarity
to category $c$ at segment $t$, and is otherwise limited by the lower of
the two modality scores. Equivalently, the lower of the two modality
scores is unchanged, while the higher score is replaced by the average of
the two scores.

We also compute the temporal average of these scores for each category:
\begin{align*}
\bar{b}_{c}^{m} = \frac{1}{T}\sum_{t=1}^{T} b_{tc}^{m},
\qquad m \in \{a,v\}.
\end{align*}

We use $\mathcal{G}$ as shorthand for comparing category $c$ with the
strongest category other than $c$. For the segment-level score tables used
below, namely $\mathbf{b}^{a}$, $\mathbf{b}^{v}$, $\mathbf{s}^{a}$, and
$\mathbf{s}^{v}$, this operation is
\begin{align*}
\mathcal{G}(\mathbf{u})_{tc}
=
\max_{c' \neq c} u_{tc'} - u_{tc}.
\end{align*}
Here $\mathbf{u}$ denotes whichever one of these score tables is being
used.
For the clip-level averaged scores, we apply the same category comparison
directly to $\bar{\mathbf{b}}^{a}$ and $\bar{\mathbf{b}}^{v}$:
\begin{align*}
\mathcal{G}(\bar{\mathbf{b}}^{m})_{c}
=
\max_{c' \neq c} \bar{b}_{c'}^{m} - \bar{b}_{c}^{m},
\qquad m \in \{a,v\}.
\end{align*}
The value is positive when category $c$ is below some competing category,
and negative when category $c$ is higher than all competing categories.

We also define the difference between the current modality-specific
similarity score and the corresponding clip-level averaged score:
\begin{align*}
\beta_{tc}^{a} &= s_{tc}^{a} - \bar{b}_{c}^{a}, \\
\beta_{tc}^{v} &= s_{tc}^{v} - \bar{b}_{c}^{v}.
\end{align*}

The audio and visual thresholds are (restated from Eqs.~(\ref{eq:tau_a})--(\ref{eq:tau_v}))
\begin{align*}
\tau_{tc}^{a}
&= \theta_0
 + \theta_1 \mathcal{G}(\mathbf{s}^{v})_{tc}
 + \theta_2 \mathcal{G}(\mathbf{b}^{a})_{tc}
 + \theta_3 \mathcal{G}(\bar{\mathbf{b}}^{a})_c
 + \theta_4 \beta_{tc}^{a}, \\
\tau_{tc}^{v}
&= \theta_5
 + \theta_6 \mathcal{G}(\mathbf{s}^{a})_{tc}
 + \theta_7 \mathcal{G}(\mathbf{b}^{v})_{tc}
 + \theta_8 \mathcal{G}(\bar{\mathbf{b}}^{v})_c
 + \theta_9 \beta_{tc}^{v}.
\end{align*}

Table~\ref{tab:fform_ranges} lists the parameter ranges assigned by the
Design Agent; the corresponding code is shown in Fig.~\ref{fig:designform}.

\begin{table}[h]
\centering
\caption{Parameter ranges for the discovered formulation. Indices are renumbered relative to the code in Fig.~\ref{fig:designform} so that audio and visual terms align ($\theta_k \leftrightarrow \theta_{k+5}$); the code index gives the position in \texttt{params}.}
\label{tab:fform_ranges}
\small
\begin{tabular}{clcc}
\toprule
Parameter & Multiplied term & Range & Code index \\
\midrule
$\theta_0$ & audio bias & $[-0.1,\,0.5]$ & \texttt{params[0]} \\
$\theta_1$ & $\mathcal{G}(\mathbf{s}^{v})_{tc}$ & $[0.0,\,2.0]$ & \texttt{params[2]} \\
$\theta_2$ & $\mathcal{G}(\mathbf{b}^{a})_{tc}$ & $[0.0,\,2.0]$ & \texttt{params[1]} \\
$\theta_3$ & $\mathcal{G}(\bar{\mathbf{b}}^{a})_c$ & $[0.0,\,1.8]$ & \texttt{params[3]} \\
$\theta_4$ & $\beta_{tc}^{a}$ & $[0.0,\,1.8]$ & \texttt{params[4]} \\
$\theta_5$ & visual bias & $[0.0,\,0.8]$ & \texttt{params[5]} \\
$\theta_6$ & $\mathcal{G}(\mathbf{s}^{a})_{tc}$ & $[0.0,\,1.8]$ & \texttt{params[9]} \\
$\theta_7$ & $\mathcal{G}(\mathbf{b}^{v})_{tc}$ & $[0.0,\,2.1]$ & \texttt{params[6]} \\
$\theta_8$ & $\mathcal{G}(\bar{\mathbf{b}}^{v})_c$ & $[0.0,\,2.0]$ & \texttt{params[7]} \\
$\theta_9$ & $\beta_{tc}^{v}$ & $[0.0,\,2.0]$ & \texttt{params[8]} \\
\bottomrule
\end{tabular}
\end{table}

Each threshold consists of a bias and four score-dependent quantities.
The terms $\mathcal{G}(\mathbf{s}^{v})_{tc}$ in the audio threshold and
$\mathcal{G}(\mathbf{s}^{a})_{tc}$ in the visual threshold use the
opposite modality's category scores at the same segment. If the opposite
modality scores category $c$ above its competitors, the quantity is
negative and the threshold is reduced. If the opposite modality scores
another category higher, the threshold is increased.

The terms $\mathcal{G}(\mathbf{b}^{a})_{tc}$ and
$\mathcal{G}(\mathbf{b}^{v})_{tc}$ compare category $c$ with the strongest
competing category at the same segment after the modality score has been
averaged with $q_{tc}$. The Design Agent assigned non-negative ranges to
the corresponding coefficients, so this part raises the threshold when a
competing category is larger under these averaged scores and lowers the
threshold when category $c$ is larger.

The terms $\mathcal{G}(\bar{\mathbf{b}}^{a})_c$ and
$\mathcal{G}(\bar{\mathbf{b}}^{v})_c$ apply the same category comparison
after averaging $b_{tc}^{m}$ over all temporal segments of the video.
This part therefore adjusts the threshold using category competition at
the clip level rather than only at the current segment.

Finally, $\beta_{tc}^{a}$ and $\beta_{tc}^{v}$ compare the current
modality-specific similarity score with the corresponding score averaged
over all temporal segments of the video, $\bar{b}_{c}^{m}$. Since the
Design Agent also assigned non-negative ranges to these coefficients,
this part raises the threshold when the current modality-specific score
is above this temporal average, and lowers it when the current score is
below it.

\paragraph{Interpretation by cue group.}
Table~\ref{tab:cue_groups} groups the eight weighted terms of $\tau^{a}_{tc}$ and $\tau^{v}_{tc}$ into four cue groups, each appearing symmetrically in the audio and visual thresholds. Taken together, $F^\star$ implements a simple principle: accept a category when both modalities support it relative to competing categories, both locally and over the full video. The policy predicts $\boldsymbol{\theta}_{0:9}$ per video, controlling the strength of each cue.

\begin{table}[h]
\centering
\caption{Cue groups in the discovered formulation $F^\star$.}
\label{tab:cue_groups}
\small
\renewcommand{\arraystretch}{1.15}
\begin{tabularx}{\textwidth}{@{}l l X@{}}
\toprule
Cue group & Terms & Interpretation \\
\midrule
Cross-modal support & $\theta_1\mathcal{G}(\mathbf{s}^{v})_{tc}$, $\theta_6\mathcal{G}(\mathbf{s}^{a})_{tc}$ & Each modality's threshold uses the other modality's category ranking at the same segment. \\
Local AV agreement & $\theta_2\mathcal{G}(\mathbf{b}^{a})_{tc}$, $\theta_7\mathcal{G}(\mathbf{b}^{v})_{tc}$ & $\mathbf{b}^{m}=\tfrac{1}{2}(\mathbf{s}^{m}+\min(\mathbf{s}^{a},\mathbf{s}^{v}))$ favors categories supported by both modalities. \\
Clip-level competition & $\theta_3\mathcal{G}(\bar{\mathbf{b}}^{a})_{c}$, $\theta_8\mathcal{G}(\bar{\mathbf{b}}^{v})_{c}$ & Compares the category against its competitors over the full video. \\
Temporal calibration & $\theta_4\beta^{a}_{tc}$, $\theta_9\beta^{v}_{tc}$ & Compares the current segment with the clip-level evidence for the same category. \\
\bottomrule
\end{tabularx}
\end{table}

\paragraph{Comparison with AV$^2$A.}
AV$^2$A~\citep{av2a} thresholds a single fused audio-visual score and updates one running threshold sequentially over segments. In contrast, $F^\star$ retains separate audio and visual scores and computes two cross-coupled thresholds from the opposite modality's ranking, segment- and clip-level category competition, and temporal deviation. The search therefore discovers a different threshold-generating equation rather than merely different threshold values.

\paragraph{Cue-group ablation.}
To identify which cues matter most, we remove each cue group from $F^\star$ in turn and retrain and evaluate the policy on the reduced formulation. Removing the cross-modal support terms causes the largest drop, from $60.2$ to $55.0$ Total Avg., identifying cross-modal support as the most important component of the discovered rule.

\section{Transferability of the Discovered Formulation}
\label{app:transfer}

A natural question is whether a discovered formulation can be reused when the experimental setting changes. This section evaluates reusing the final formulation $F^\star$ (Appendix~\ref{app:fform}) without any new search: we keep $F^\star$ fixed and retrain only the lightweight policy network for each new setting, using the same policy-training and validation-selection protocol as the main experiments. Retraining runs on cached frozen embeddings, takes approximately 12 minutes, and requires no LLM calls or oracle search. We consider two axes of transfer: replacing the frozen encoder and changing the task.

\paragraph{Cross-encoder transfer.}
We first test whether $F^\star$, discovered with ImageBind, remains effective when the frozen encoder is replaced. Table~\ref{tab:enc_transfer} compares the transferred $F^\star$ against AV$^2$A re-evaluated on OV-AVEBench with its published per-encoder hyperparameters. With LanguageBind, the transferred $F^\star$ reaches $58.6$ Total Avg., within $1.6$ points of the ImageBind result ($60.2$) and $+8.0$ over AV$^2$A. With CLIP+CLAP, both methods degrade, yet the transferred $F^\star$ still outperforms AV$^2$A by $+5.6$. Thus, the discovered rule is not specific to ImageBind, and changing the encoder requires only policy retraining rather than repeating the formulation search.

\begin{table}[h]
\centering
\caption{Cross-encoder transfer on OV-AVEBench (Total Avg.). AV$^2$A uses its published per-encoder hyperparameters; the transferred $F^\star$ keeps the ImageBind-discovered formulation and retrains only the policy.}
\label{tab:enc_transfer}
\small
\begin{tabular}{lcc}
\toprule
Encoder & AV$^2$A & Transferred $F^\star$ \\
\midrule
LanguageBind & 50.6 & \textbf{58.6} \\
CLIP+CLAP & 48.9 & \textbf{54.5} \\
\bottomrule
\end{tabular}
\end{table}

\paragraph{Cross-task transfer.}
We next transfer $F^\star$ to the two open-vocabulary tasks of Section~\ref{sec:generalization}, following the protocols in Appendix~\ref{app:gen}. The task-specific auxiliary formulations discovered there---the event-confidence formulation for OV-DAVEL, and the pseudo-label formulation and confidence filter for OV-AVVP---are kept fixed; only the threshold formulation is replaced by $F^\star$, after which the policy is retrained and the auxiliary parameters are re-fitted. Table~\ref{tab:transfer} compares the transferred $F^\star$ with task-specific discovery. On OV-DAVEL (50:50), the transferred formulation exceeds the baseline by $10.9$ mAP and falls only $2.4$ points below task-specific discovery, remaining competitive across localization settings. Transfer to OV-AVVP is partial, as the task requires weakly supervised, modality-wise multi-label parsing rather than jointly supported AV-event localization; note that the transferred $F^\star$ was discovered for a different output structure, whereas the AV$^2$A rule was designed with AVVP in view. Thus, $F^\star$ is reusable across closely related localization settings, while structurally different supervision and output spaces benefit from task-specific rediscovery.

\begin{table}[h]
\centering
\caption{Cross-task transfer: task-specific discovery versus transferring the fixed OV-AVEBench formulation $F^\star$ with the policy retrained and the task-specific auxiliary parameters re-fitted.}
\label{tab:transfer}
\small
\begin{tabular}{llccc}
\toprule
Benchmark & Metric & Baseline & Task-specific & Transferred $F^\star$ \\
\midrule
OV-DAVEL (50:50) & mAP & 19.4 & \textbf{32.7} & 30.3 \\
OV-AVVP (17:8) & Seg.\ Type@AV & 52.4 & \textbf{55.6} & 49.5 \\
\bottomrule
\end{tabular}
\end{table}

\section{Sensitivity Analyses}
\label{app:sens}

This section reports additional sensitivity analyses of the search procedure and the learned policy. Unless noted otherwise, each setting follows the main protocol: three isolated 20-iteration sessions, with mean $\pm$ std.\ over the three session-level validation-selected candidates. The main-paper score of 60.2 is the final candidate selected across sessions by validation performance.

\paragraph{LLM backbone.}
We change only the LLM backbone used by all four agents, keeping the prompts, formulation interface, search budget, policy training, and selection unchanged (Table~\ref{tab:backbone}). All five backbones produce valid executable formulations, though the results appear to be influenced by the backbone's reasoning capability. GPT-5.6 Sol, released after our main experiments, performs comparably to GPT-5.4 (59.5 vs.\ 59.4, within one standard deviation); all main results use GPT-5.4.

\begin{table}[h]
\centering
\caption{Sensitivity to the LLM backbone (Test Avg., mean $\pm$ std.\ over three sessions).}
\label{tab:backbone}
\small
\begin{tabular}{lcc}
\toprule
Backbone & Access & Test Avg. \\
\midrule
GPT-5.6 Sol & Closed & \textbf{59.5 $\pm$ 0.9} \\
GPT-5.4 & Closed & 59.4 $\pm$ 0.9 \\
Claude Opus 4.8 & Closed & 57.4 $\pm$ 1.1 \\
Qwen3.7-Max & Closed & 57.1 $\pm$ 1.8 \\
GLM-5.2 & Open-weight & 56.7 $\pm$ 1.0 \\
\bottomrule
\end{tabular}
\end{table}

\paragraph{Design constraints.}
We independently relax each formal constraint of the Design Agent (Section~\ref{sec:design_agent}): the linear-form constraint, the predefined constant set, or the per-function statement budget. Each variant removes one Design-Agent instruction and its corresponding AST check, keeping $P{=}10$, the LLM, the search budget, policy training, and selection unchanged. As shown in Table~\ref{tab:constraints}, every relaxation reduces validation and test performance. Removing linearity leaves oracle capacity nearly unchanged but reduces Val Acc.\ by 2.6 and Test Avg.\ by 4.3, indicating harder policy instantiation. The other relaxations increase formulation complexity without benefit---up to 26 numeric constants, or $1.5\times$ more primitives and $1.3\times$ more statements---with larger oracle variance. The constraints therefore act as structural regularizers for realizability and stable search.

\begin{table}[h]
\centering
\caption{Sensitivity to the Design Agent's formal constraints (mean $\pm$ std.\ over three sessions).}
\label{tab:constraints}
\small
\begin{tabular}{lccc}
\toprule
Setting & Oracle Acc. & Val Acc. & Test Avg. \\
\midrule
CueRator (default) & \textbf{88.7 $\pm$ 0.9} & \textbf{67.8 $\pm$ 0.2} & \textbf{59.4 $\pm$ 0.9} \\
w/o linear form & 88.4 $\pm$ 0.3 & 65.2 $\pm$ 0.6 & 55.1 $\pm$ 0.8 \\
w/o constant set & 85.6 $\pm$ 4.1 & 66.4 $\pm$ 0.8 & 56.6 $\pm$ 1.7 \\
w/o statement budget & 86.6 $\pm$ 5.6 & 66.6 $\pm$ 0.5 & 57.0 $\pm$ 0.2 \\
\bottomrule
\end{tabular}
\end{table}

\paragraph{Number of formulation parameters.}
Under the constrained interface, each modality-specific threshold contains one bias and four weighted symbolic terms, giving $P = 2\times(1+4) = 10$ parameters. This budget was fixed before evaluation and retained across all main experiments; it was not selected using test performance. Table~\ref{tab:num_params} varies only the number of weighted terms per modality under the same search and policy protocol. Performance decreases with both smaller and larger budgets, suggesting that smaller $P$ limits expressiveness, whereas larger $P$ makes formulation search and policy instantiation less stable. Search time grows only moderately with $P$.

\begin{table}[h]
\centering
\caption{Sensitivity to the number of formulation parameters $P$ (mean $\pm$ std.\ over three sessions).}
\label{tab:num_params}
\small
\begin{tabular}{cccc}
\toprule
$P$ & Terms per modality (incl.\ bias) & Search time (h/session) & Test Avg. \\
\midrule
6 & 3 & 8.2 $\pm$ 0.2 & 54.6 $\pm$ 0.3 \\
8 & 4 & 9.0 $\pm$ 0.2 & 58.3 $\pm$ 1.1 \\
\textbf{10} & \textbf{5} & 9.8 $\pm$ 0.2 & \textbf{59.4 $\pm$ 0.9} \\
12 & 6 & 10.0 $\pm$ 0.1 & 57.2 $\pm$ 1.0 \\
14 & 7 & 10.6 $\pm$ 0.4 & 55.3 $\pm$ 1.9 \\
\bottomrule
\end{tabular}
\end{table}

\paragraph{Policy reward.}
The policy is trained with per-video segment accuracy as the reward (Appendix~\ref{app:poll}), while the final evaluation also includes segment-level and event-level F1. To test whether a reward closer to the final metrics helps, we retrain the policy for the selected formulation $F^\star$ using the average of Acc., Seg., and Eve.\ as the reward (Table~\ref{tab:reward}). The metric-aligned reward provides a modest improvement in event-level F1 (52.7 $\rightarrow$ 53.0) but slightly reduces accuracy and the overall average (60.2 $\rightarrow$ 59.8). We therefore retain per-video accuracy as the default reward, which achieves the best overall performance while requiring approximately $2\times$ less policy-training time.

\begin{table}[h]
\centering
\caption{Effect of the policy reward on the selected formulation $F^\star$ (OV-AVEBench test, Total).}
\label{tab:reward}
\small
\begin{tabular}{lcccc}
\toprule
Reward & Acc. & Seg. & Eve. & Avg. \\
\midrule
Acc.\ (default) & \textbf{68.6} & \textbf{59.4} & 52.7 & \textbf{60.2} \\
Metric average & 67.2 & 59.2 & \textbf{53.0} & 59.8 \\
\bottomrule
\end{tabular}
\end{table}

\section{Qualitative Analysis}
\label{app:qual}

\begin{figure*}[!t]
\centering
\includegraphics[width=\textwidth]{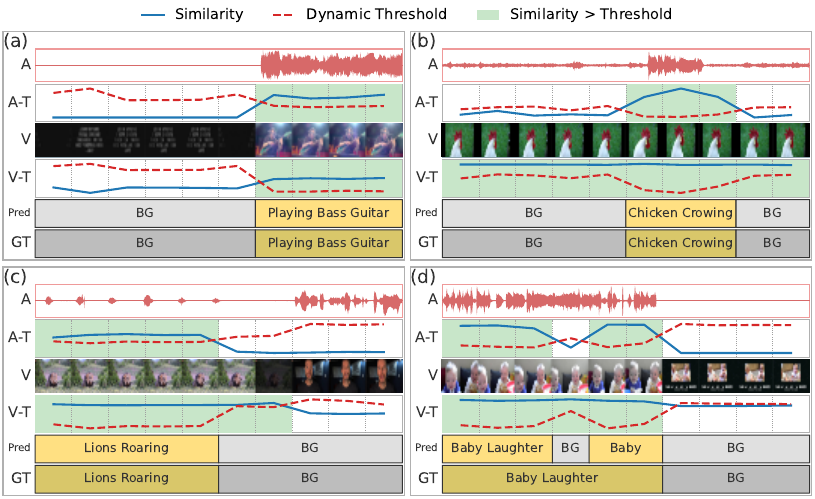}
\caption{\textbf{Per-video adaptive thresholds across diverse event types.} Each panel shows audio (A) and visual (V) signals, their similarities to the target category (A-T, V-T), and CueRator's per-segment dynamic thresholds (red dashed). Green regions denote similarity > threshold; Pred and GT bars below give predicted and ground-truth labels.}
\label{fig:qualitative}
\end{figure*}

Fig.~\ref{fig:qualitative} illustrates CueRator's per-segment adaptation. In~(a), audio and visual co-occur, and the threshold drops cleanly over the event. In~(b), the visual is constant while the audio is sparse; the threshold lowers within audio segments and rises elsewhere, cleanly identifying the event window. In~(c), the audio fades before the visual: both thresholds rise gradually, with A-T similarity crossing below its threshold first to mark the audio side as background; once the visual also fades, V-T's threshold rises further to firmly partition the visual side as well. Panel~(d) is a failure: when A-T similarity drops within the event due to encoder noise, the threshold rises instead of compensating. Across panels, threshold movements in one modality often track the other, reflecting that supervision is provided only at the AV-event level.

To complement Fig.~\ref{fig:qualitative}, we provide twelve further successful predictions covering temporal patterns not represented in the main figure: post-event modality fade, persistent visual with sparse audio, mid-segment modality switch, late-onset audio, single-pulse audio, multi-event sequences with short gaps, full-clip continuous events, and correct background classification including AV-agreement rejection of single-modality activations. Across all panels, CueRator's per-segment thresholds adjust to each modality's local temporal signature, and the AV-agreement rule resolves modality disagreements consistently with the ground truth. \emph{Conversely}, we also illustrate four failure modes that surface during qualitative inspection: a missed within-event background gap, delayed event boundaries, semantically close category confusion, and under-detection of repeated short events.

\begin{figure}[h]
  \centering
  \includegraphics[width=\textwidth]{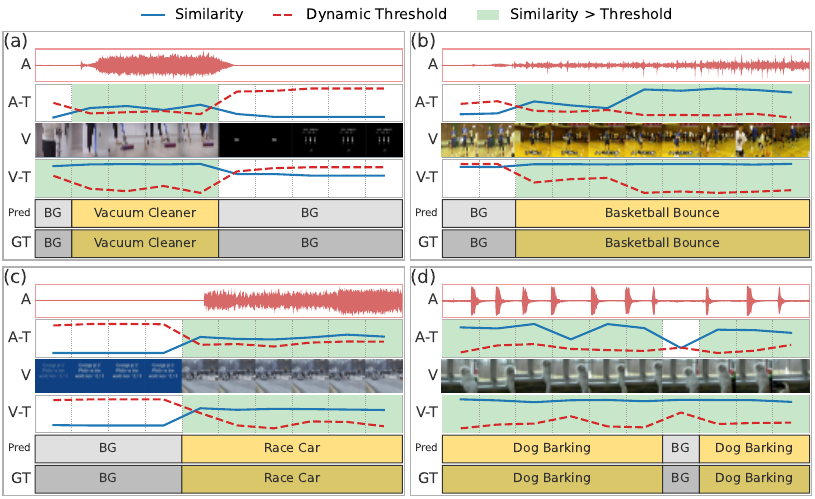}
  \caption{\textbf{Additional qualitative results (set 1).}
  (a)~\textit{Vacuum Cleaner}: both modalities fade after the event; thresholds rise to mark the post-event background.
  (b)~\textit{Basketball Bounce}: visual stays consistent while audio builds up from silence; the audio threshold lowers accordingly, capturing the onset.
  (c)~\textit{Race Car}: the visual switches to the target as audio fades in mid-clip; both thresholds drop, marking the event window.
  (d)~\textit{Dog Barking}: a brief silent gap within the event is detected as background as the audio threshold rises above A-T there.}
  \label{fig:qualitative_app1}
\end{figure}

\begin{figure}[h]
  \centering
  \includegraphics[width=\textwidth]{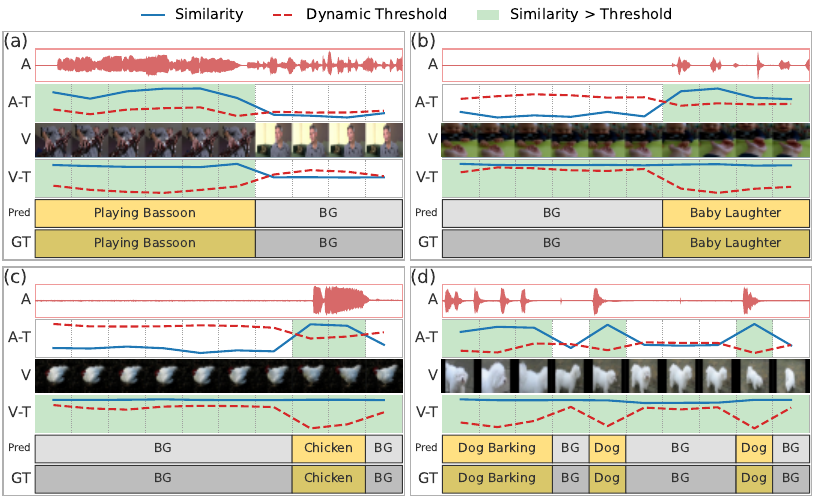}
  \caption{\textbf{Additional qualitative results (set 2).}
  (a)~\textit{Playing Bassoon}: strong audio at start fades out together with the visual; thresholds rise jointly afterward.
  (b)~\textit{Baby Laughter}: audio onset is delayed; thresholds drop only when both modalities support the event.
  (c)~\textit{Chicken}: a brief late audio pulse aligns with a target-consistent visual; the threshold isolates the short event window precisely.
  (d)~\textit{Dog Barking}: multiple disjoint events with a brief background gap; thresholds track audio pulses while the persistent visual stays above its threshold.}
  \label{fig:qualitative_app2}
\end{figure}

\begin{figure}[h]
  \centering
  \includegraphics[width=\textwidth]{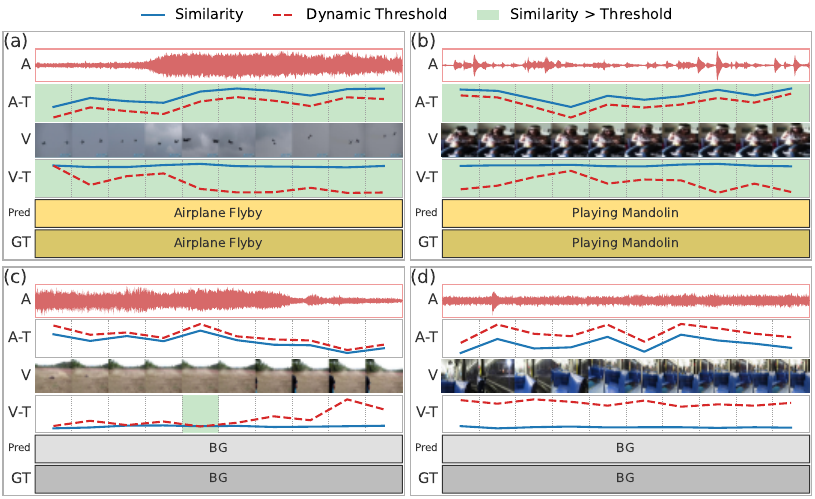}
  \caption{\textbf{Additional qualitative results (set 3).}
  (a)~\textit{Airplane Flyby}: a full-clip event with gradual audio buildup; both thresholds stay below A-T and V-T throughout, sustaining continuous detection.
  (b)~\textit{Playing Mandolin}: a stable full-clip event with clean cues; both similarities remain above their thresholds throughout, with no spurious flips.
  (c)~\textit{Background}: a brief V-T spike does not trigger a false positive; A-T stays below its threshold, so the AV-agreement rule rejects the segment.
  (d)~\textit{Background}: both A-T and V-T stay below their thresholds throughout, keeping the entire clip correctly classified as background.}
  \label{fig:qualitative_app3}
\end{figure}

\begin{figure}[h]
  \centering
  \includegraphics[width=\textwidth]{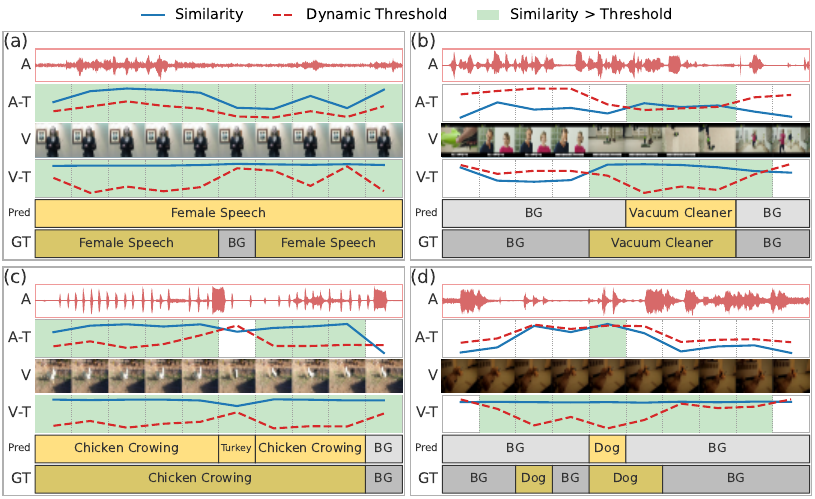}
  \caption{\textbf{Failure cases.}
  (a)~\textit{Female Speech}: a brief background gap within the speech is missed; the audio threshold tracks the dipping A-T instead of rising, so neither modality flips.
  (b)~\textit{Vacuum Cleaner}: the event window is detected with delayed onset and early offset; both thresholds drop only once the audio is fully established.
  (c)~\textit{Chicken Crowing}: a mid-clip segment is briefly misclassified as a semantically close category (\textit{Turkey}); the encoder's similarity favors the wrong class within that window.
  (d)~\textit{Dog Barking}: a few boundary segments around the event are missed; the audio threshold tracks A-T closely and crosses above it at minor dips.}
  \label{fig:qualitative_failure}
\end{figure}

\end{document}